\documentclass[10pt,twocolumn]{article}

\usepackage[letterpaper,margin=0.75in,columnsep=0.28in]{geometry}
\usepackage[T1]{fontenc}
\usepackage[utf8]{inputenc}
\usepackage{mathptmx}
\usepackage{graphicx}
\usepackage{booktabs}
\usepackage{array}
\usepackage{calc}
\usepackage{caption}
\usepackage{hanging}
\usepackage{microtype}
\usepackage{enumitem}
\setlist{nosep,leftmargin=1.4em}
\usepackage[hidelinks]{hyperref}
\usepackage{titlesec}
\titleformat{\section}{\large\bfseries}{\thesection.}{0.5em}{}
\titleformat{\subsection}{\normalsize\bfseries}{\thesubsection}{0.5em}{}
\titleformat{\subsubsection}{\normalsize\bfseries\itshape}{\thesubsubsection}{0.5em}{}
\titlespacing*{\section}{0pt}{10pt}{4pt}
\titlespacing*{\subsection}{0pt}{8pt}{3pt}
\titlespacing*{\subsubsection}{0pt}{6pt}{2pt}
\title{\LARGE\bfseries Measuring How Students Rely on Generative AI in Academic Writing: Development and Multi-Source Validation of the Generative AI Reliance Types Scale (GenAI-RTS)}

\author{%
\small
\begin{tabular}{c}
\normalsize\textbf{Shahin Hossain}\\[2pt]
School of Education\\
University of Maryland Baltimore County\\
Baltimore, Maryland, USA\\
{\small \href{mailto:shahinh1@umbc.edu}{shahinh1@umbc.edu}}\\
{\small ORCID: \href{https://orcid.org/0000-0002-3461-1147}{0000-0002-3461-1147}}
\end{tabular}%
\hspace{1.5em}%
\begin{tabular}{c}
\normalsize\textbf{Tukhbita Afroz Nawmi}\\[2pt]
Department of Counseling, School and\\
Educational Psychology, Graduate School of Education\\
University at Buffalo, Buffalo, New York, USA\\
{\small \href{mailto:tukhbita@buffalo.edu}{tukhbita@buffalo.edu}}\\
{\small ORCID: \href{https://orcid.org/0009-0003-4918-1420}{0009-0003-4918-1420}}
\end{tabular}%
}
\date{}

\begin{document}

\twocolumn[
\maketitle
\vspace{-18pt}
]

\begin{abstract}
\noindent As generative AI (GenAI) becomes increasingly embedded in undergraduate academic writing, how students rely on these tools, rather than simply whether they use them, has become a central question for learning, academic integrity, and educational equity. Existing measures of reliance were developed inductively, focused on discrete problem-solving tasks, and validated mainly with homogeneous samples. The field lacks a theory-driven instrument for extended writing contexts. This study developed and validated the GenAI Reliance Types Scale (GenAI-RTS), a 20-item instrument measuring four theoretically derived types of GenAI reliance: Strategic, Instrumental, Dependent, and Dialogic. Validation followed the multisource framework of \emph{Standards for Educational and Psychological Testing}, drawing on a survey of 382 undergraduates at a U.S. Minority-Serving Institution and interviews with 14 purposively sampled students. Confirmatory factor analyses of six competing models indicated support for a five-factor structure. Within this structure, Strategic Reliance consists of two facets, Deliberate Use and Critical Evaluation, in addition to Instrumental, Dependent, and Dialogic factors (CFI = .92, RMSEA = .08; DWLS CFI = .98, RMSEA = .07). Subscale reliability was acceptable to good ($\omega$ = .75--.88), and scalar measurement invariance held across gender, first-generation status, and STEM/non-STEM majors; to our knowledge, the first such evidence for an instrument measuring GenAI reliance. Rasch analysis indicated that a five-point response format would improve category functioning.~Strategic reliance was positively associated with AI literacy; Critical Evaluation and Dependent reliance emerged as empirically distinct constructs; and the reliance types differentiated students across multiple writing process and outcome variables. The GenAI-RTS offers researchers and educators a theoretically grounded, psychometrically validated instrument for identifying undergraduate reliance profiles and supporting research, assessment, and AI literacy intervention.
\end{abstract}

\vspace{4pt}
\noindent\textbf{Keywords:} Generative artificial intelligence; ChatGPT; AI Literacy; Measurement invariance; Scale Development; Higher Education

\vspace{8pt}

\textbf{1. Introduction}

When ChatGPT reached an estimated 100 million users within two months of
its November 2022 release, making it, at the time, the "fastest-growing
consumer application in history" (Hu, 2023, para. 1), it entered higher
education at unprecedented speed. Adoption was primarily student-driven,
outpacing institutional policy development, faculty preparation, and
empirical research: reviewing the field only weeks after release,
Rudolph et al. (2023) found a "dearth of empirical research" (p. 352) on
the tool\textquotesingle s educational implications even as institutions
scrambled to formulate policy responses (Cotton et al., 2024). Use has
since approached universality: among UK undergraduates surveyed
annually, the proportion using AI in at least one way rose from 66\% in
2024 to 92\% in 2025 (Freeman, 2025) and 95\% in 2026, with 94\% now
using generative AI (GenAI) to support assessed work (Stephenson \&
Armstrong, 2026). Academic writing is a key domain of this use: in the
largest early study of adopters, content creation and editing,
composing, summarizing, and proofreading text accounted for 78.11\% of
reported ChatGPT applications in higher education (Mogavi et al., 2024).
Students entering college since Fall 2022 are thus the first cohort
whose undergraduate writing development has unfolded entirely within a
GenAI-rich environment. Throughout this article, GenAI denotes the large
language model (LLM)-based text-generation tools that students
overwhelmingly use in academic writing; because LLMs are the technology
underlying these tools, the theoretical analysis draws on both the GenAI
and human--AI reliance literatures.

Under these conditions, the question that has guided most institutional
policy and research,~\emph{whether}~students use GenAI, has lost its
diagnostic value. When use is nearly universal, the distinction between
use and non-use no longer differentiates students meaningfully: it
cannot distinguish a student who critically evaluates and selectively
integrates AI-generated content from one who accepts and submits it with
little or no evaluation. Yet the difference between these two students
is precisely what matters for learning, integrity, and equity. For
learning, writing develops thinking through cognitive struggle (Bereiter
\& Scardamalia, 1987; Kellogg, 2008), and emerging evidence points in a
consistent direction at two levels of specificity: in a four-month EEG
study of essay writing, participants composing with LLM assistance
showed the weakest brain connectivity of three tool conditions and
underperformed at neural, linguistic, and behavioral levels (Kosmyna et
al., 2025), while survey evidence links frequent AI-tool use to lower
critical-thinking scores, a relationship mediated by cognitive
offloading (Gerlich, 2025); both the writing literature and this newer
evidence indicate that the effects depend on how the tool is engaged,
not merely whether it is used. For integrity, current frameworks rely on
the same binary, treating qualitatively different forms of GenAI
engagement as a single category of misconduct, even as nearly all
students, 94\%, now use GenAI to support assessed work (Stephenson \&
Armstrong, 2026). For equity, access to AI technologies is unevenly
distributed and risks widening existing educational inequalities
(Bulathwela et al., 2024), and a socioeconomic divide in student AI use
is already documented (Freeman, 2025); whether students from
under-resourced backgrounds also develop less productive patterns of
reliance is a question only an instrument measuring qualitatively
distinct reliance patterns can answer. The pressing empirical question
is therefore not whether students rely on GenAI but how.

Answering that question requires a new instrument, for three reasons.
First, existing measures were developed for discrete decision-making
tasks rather than extended academic writing. Second, the most closely
related scales measure adjacent constructs or were developed inductively
rather than from theory. Third, none provides evidence of scalar
measurement invariance across demographic groups, the evidence needed to
support meaningful equity-related comparisons.

First, the reliance construct was developed largely within
decision-support paradigms, where trust is conceptualized as an attitude
and reliance as the behavior it guides (Lee \& See, 2004), and where
reliance is operationalized through discrete, outcome-linked behaviors:
adjusting a judgment toward advice received (Logg et al., 2019) or
accepting versus overriding an algorithm\textquotesingle s
recommendation, with appropriateness defined by the
advice\textquotesingle s correctness (Schemmer et al., 2023). These
operationalizations presuppose tasks with identifiable decision points
and verifiable outcomes. Hou et al. (2025) advanced the field on
precisely these grounds, arguing that outcome-linked measures are
unsuited to complex problem-solving tasks whose outcomes are neither
binary nor immediately clear, and developing a self-report scale that
measures reliance behaviors without linking them to outcomes. Academic
writing extends this logic further still: it is extended, iterative, and
evaluatively open-ended, with no correctness signal against which
reliance could be scored as appropriate or excessive, and it is among
the most common and educationally consequential applications of GenAI in
higher education, as the evidence on adoption above indicates. The
present study extends reliance measurement to this domain.

Second, the task-domain gap is compounded by a construct and method gap:
the rapidly growing family of related instruments measures adjacent
constructs through predominantly inductive methods. Recent scales
capture AI dependence as a maladaptive disposition encompassing
emotional dependence and loss of control (Wu et al., 2026), student
learning agency in GenAI-supported contexts (Xia et al., 2025), and
students\textquotesingle{} general use of GenAI tools (Barcelona \& Dela
Cruz, 2025). Reliance itself, the behavioral repertoire through which
students engage GenAI during authentic academic tasks, remains measured
by a single validated instrument (Hou et al., 2025), whose four factors
(reflective, cautious, thoughtless, and collaborative use) emerged from
exploratory factor analysis and were named post hoc. Inductive
approaches are valuable for mapping new phenomena, but the resulting
factors remain tied to the items, tasks, and samples from which they
were derived, and even where dimensions have been hypothesized in
advance (Wu et al., 2026), the retained structures were established
through exploratory extraction rather than tested against rival
theoretical specifications. A fully deductive approach, specifying a
typology from theory, operationalizing it a priori, and testing it
against competing structures has not, to our knowledge, been undertaken
for GenAI reliance. The present study derives four reliance types
(Strategic, Instrumental, Dependent, and Dialogic) from established
frameworks in metacognition (Schraw \& Dennison, 1994), critical AI
literacy (Long \& Magerko, 2020), and writing research; Section 2.3
develops each in full.

Third, existing validation studies are limited in both the diversity of
their samples and the breadth of validity evidence they report. Most
have been conducted within a single institution or national context: Hou
et al. (2025) acknowledge that their scale was validated with "a
relatively homogeneous sample of undergraduate students from a single
Asian university" (p. 10), and the same constraint characterizes newer
AI dependence and GenAI use scales (Barcelona \& Dela Cruz, 2025; Wu et
al., 2026). Where measurement invariance has been examined at all,
testing has been limited to configural and metric checks across split
validation samples (Wu et al., 2026) rather than to scalar invariance
across demographic groups, as required for meaningful latent mean
comparisons. It therefore remains unclear whether these instruments
function equivalently across diverse student populations. Validation
evidence has also typically been reported as a collection of statistical
indices rather than as a coherent validity argument: the~\emph{Standards
for Educational and Psychological Testing}~(American Educational
Research Association {[}AERA{]}, American Psychological Association
{[}APA{]}, \& National Council on Measurement in Education {[}NCME{]},
2014) call for multiple complementary sources of validity evidence,
integrated on the argument-based view of validation into a structured
case for score interpretation (Kane, 2013). No GenAI reliance instrument
has, to our knowledge, been validated within this framework or examined
for scalar invariance across demographic groups, limiting confidence in
group comparisons and equity-related interpretations.

This study addresses these three gaps by developing and validating the
GenAI Reliance Types Scale (GenAI-RTS), a 20-item self-report instrument
administered to undergraduates at a U.S. public R1 Minority-Serving
Institution. The study makes four contributions. First, to our
knowledge, it provides the first theory-derived, confirmatory-tested
typology of GenAI reliance in academic writing, evaluating the
hypothesized structure against rival one-, two-, three-, four-, and
five-factor and higher-order specifications rather than assuming it.
Second, it presents a multisource validity argument structured by
the~\emph{Standards}~(AERA et al., 2014), delivering evidence on four of
the five sources, test content, response processes, internal structure,
and relationships to other variables, including response-process
evidence gathered across the full range of scale scores, evidence that
prior studies have confined to item development or to high-scoring
respondents. Third, to our knowledge, it reports the first evidence of
scalar measurement invariance for an instrument measuring GenAI
reliance, testing configural, metric, and scalar invariance across
gender, first-generation status, and STEM versus non-STEM majors in a
demographically diverse sample in which no racial group constitutes a
majority. Fourth, it cross-validates the internal structure across two
analytic traditions, confirmatory factor analysis within the classical
test theory framework and structural equation modeling, and Rasch
measurement analysis, providing convergent psychometric evidence that
neither tradition alone provides.

Accordingly, this study asks whether the accumulated evidence supports
interpreting GenAI-RTS scores as measures of undergraduate
students\textquotesingle{} reliance on GenAI in academic writing.
Specifically, it evaluates evidence from four of the five sources
specified in the~\emph{Standards}: test content, response processes,
internal structure, and relationships to other variables. By developing
and validating the GenAI-RTS, the study provides researchers and
educators with a theory-derived instrument to identify distinct profiles
of GenAI reliance and to support future research on AI literacy,
academic writing, and responsible GenAI use.

\section{Literature Review}

\subsection{Reliance on GenAI: From decision support to generative
tasks}

Reliance on GenAI refers to the behavioral act of depending on a GenAI
tool\textquotesingle s output, which is distinct from trust in GenAI:
trust is the attitude that predicts but does not necessarily constitute
reliance (Lee \& See, 2004). In the decision-support tradition where the
construct was developed, the central concern is calibration:
over-reliance, or accepting automation despite errors (misuse), and
under-reliance, or underutilizing or rejecting automation (disuse)
(Parasuraman \& Riley, 1997), with the corresponding design goal of
appropriate reliance (Lee \& See, 2004), in which acceptance of the
system\textquotesingle s advice corresponds to the correctness of that
advice on a case-by-case basis (Schemmer et al., 2023). Reliance is
accordingly operationalized as advice-taking or judgment revision in
tasks with objectively verifiable answers (Logg et al., 2019), thereby
enabling acceptance behavior to be scored for appropriateness relative
to decision accuracy (Schemmer et al., 2023).

GenAI tools dissolve the conditions this machinery presupposes. When
students engage these tools during academic writing, there is no
discrete recommendation to accept or reject and no binary correctness
signal against which acceptance could be calibrated. Engagement is
extended, iterative, and distributed across the composing process, and
its object is co-produced text whose quality is evaluatively open-ended.
The consequences of concern have also shifted from immediate decision
accuracy to cumulative cognitive development. In a randomized
experiment, AI-assisted writers achieved greater essay-score improvement
than comparison groups while showing no advantage in knowledge gain or
transfer and reduced engagement in metacognitive processes such as
orientation and evaluation, a pattern Fan et al. (2025) term
metacognitive laziness. Preliminary neurocognitive evidence points in
the same direction (Kosmyna et al., 2025). Yet the effects are not
inherent to the tool: when LLM use is deliberately scaffolded within
strategy instruction, even young learners\textquotesingle{}
self-regulated learning and writing performance improve (Liu et al.,
2024). These studies converge on a moderating insight: the cognitive
costs attach not to use but to the manner of use. Reliance on GenAI is
therefore best conceptualized not as a quantity to be calibrated against
system accuracy but as a repertoire of qualitatively distinct behavioral
patterns whose consequences for learning differ. Hou et al. (2025)
operationalized this reconceptualization for collaborative
problem-solving, developing a self-report scale that measures reliance
behaviors without linking them to task outcomes. Academic writing,
extended, individual, identity-laden, and among the most common student
applications of GenAI (Stephenson \& Armstrong, 2026), is the natural
next domain, and the one in which, to our knowledge, no validated
reliance instrument yet exists.

\subsection{Existing measures of student engagement with
GenAI}

Table 1 summarizes the published instruments most relevant to measuring
how students engage with GenAI, compared across construct, development
approach, task context, sample, and the validity evidence provided. Two
boundary clarifications govern inclusion. First, the rapidly expanding
family of AI literacy and attitude scales (e.g., Chung et al., 2025;
Gümüş \& Kara, 2025; Marengo et al., 2025; Zhang et al., 2025) measures
competencies and evaluations rather than reliance behavior and is
treated here as adjacent rather than competing. Second, instruments
measuring AI dependence as a maladaptive psychological disposition,
exemplified by the AIDep-22, whose dimensions include emotional
dependence and loss of control (Wu et al., 2026), capture a
clinical-adjacent trait conceptually downstream of the behavioral
patterns measured here: a student may enact dependent reliance behaviors
without the affective entanglement that defines dependence as a
disposition, and the two constructs warrant separate measurement.

Three patterns emerge from Table 1, and together they define the gap
this study addresses. First, only one published instrument measures
reliance as a behavioral repertoire during authentic academic tasks (Hou
et al., 2025); the remainder measure general use practices, dependence
dispositions, or adjacent constructs. Second, development has been
predominantly inductive: factors are extracted from item pools and named
post hoc, an approach that maps unfamiliar terrain effectively but binds
the resulting structure to the items, tasks, and samples that produced
it and provides no mechanism for testing whether a theoretically
specified structure survives confrontation with rival models. Third, the
validity evidence delivered is structurally incomplete in consistent
ways: no instrument in the table reports scalar measurement invariance
across demographic groups, the precondition for the latent mean
comparisons that motivate much of this literature (invariance testing,
where reported at all, has been limited to configural and metric checks
across split validation samples; Wu et al., 2026); none organizes its
evidence as a structured validity argument; and response-process
evidence is absent, confined to item development (Wu et al., 2026), or
restricted to high-scoring respondents (Hou et al., 2025). Every
validation sample is drawn from a single institution or a single
national context. These are not incidental omissions; they jointly
constrain what scores from these instruments can legitimately be
interpreted to mean and for whom.

The practical consequence is concrete. With existing instruments,
educators cannot distinguish productive from unproductive reliance
patterns, identify students whose reliance profile warrants
intervention, evaluate whether AI-literacy programs
change~\emph{how}~students rely rather than merely how much, or compare
reliance across groups and institutions, comparisons that only scalar
invariance evidence can license. These are the uses a validated reliance
typology must support, and they define the evidentiary bar the present
study sets for itself.

\subsection{A theory-derived typology of GenAI reliance in academic
writing}

The four-type typology operationalized by the GenAI-RTS derives from the
intersection of three literatures: metacognition and self-regulated
writing, AI literacy, and approaches-to-learning theory. Its organizing
premise, supported by the process-level evidence reviewed above, is that
reliance patterns differ along two theoretically consequential
dimensions: the degree of cognitive governance the writer retains
(planning, monitoring, and evaluating the AI\textquotesingle s
contribution) and the scope of delegation (which writing functions, and
how much of each, are transferred to the system); Figure 1 maps the four
types on these dimensions.

\begin{figure}[t]
\centering
\includegraphics[width=\columnwidth]{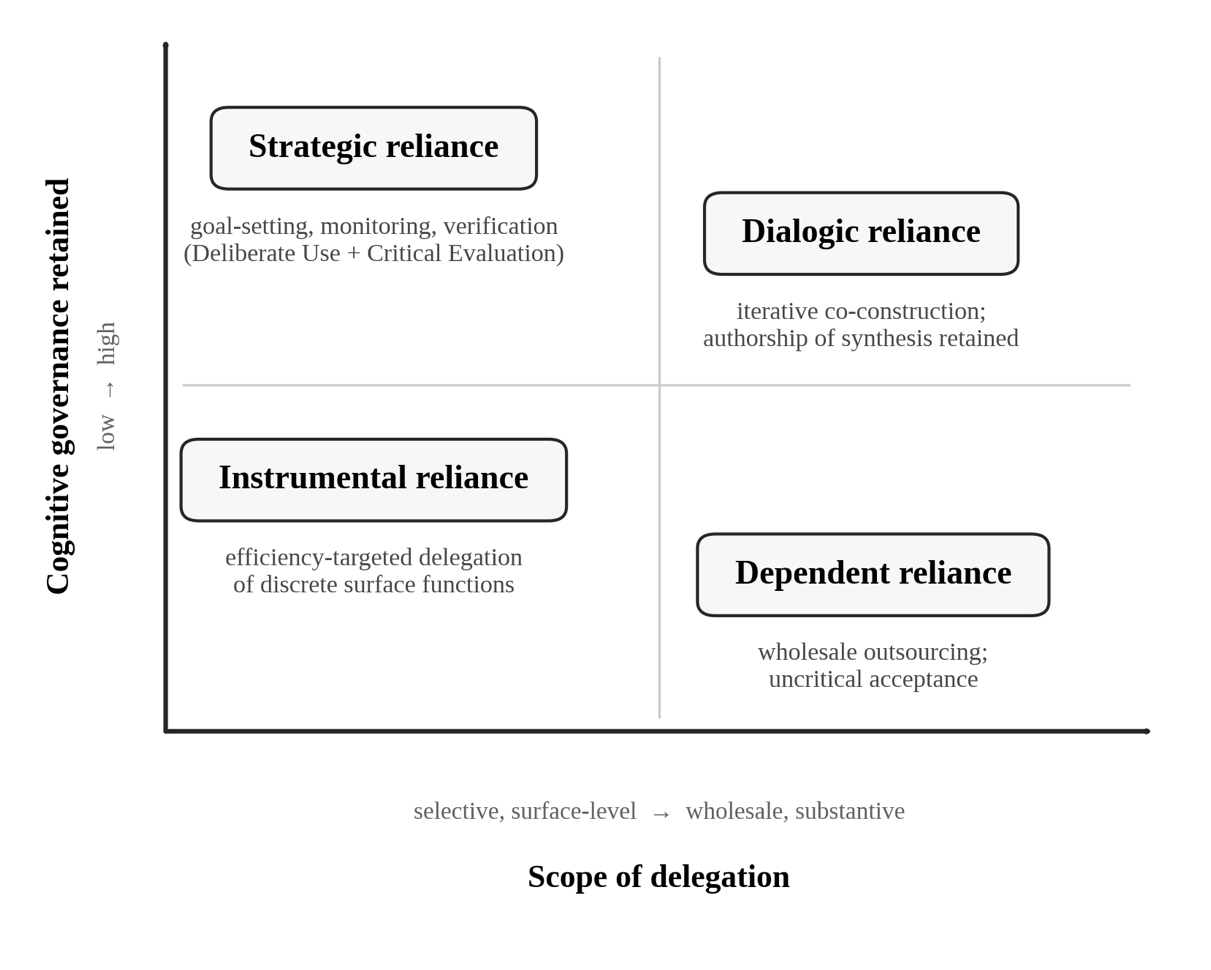}
\caption*{\textbf{Figure 1.}~\emph{Conceptual Map of the Four GenAI Reliance Types Along Two Dimensions: Cognitive Governance Retained and Scope of Delegation} \emph{\textbf{Note.}} Strategic reliance comprises the Deliberate Use and Critical Evaluation facets.}
\end{figure}

Strategic reliance denotes goal-directed engagement governed by
metacognitive orchestration and critical evaluation of output, grounded
in the regulation-of-cognition component of metacognitive theory (Schraw
\& Dennison, 1994), self-regulated writing development (Graham \&
Harris, 2000), and the evaluative competencies of AI literacy frameworks
(Long \& Magerko, 2020); these foundations imply both an orchestration
facet and an evaluative facet, and the instrument operationalizes both.
Instrumental reliance denotes efficiency-targeted delegation of
discrete, surface-level functions while the writer retains the
substantive cognitive work, anchored in the surface--deep distinction of
Biggs\textquotesingle{} (1987, 1999) approaches-to-learning framework:
the functions delegated are those a surface approach targets, while deep
engagement with content and argument is retained. Dependent reliance
denotes wholesale outsourcing across writing functions, including the
idea-generation and argument-construction work through which writing
develops thinking (Bereiter \& Scardamalia, 1987; Kellogg, 2008),
grounded in cognitive-offloading research (Lodge \& Loble, 2026; Risko
\& Gilbert, 2016) and the experimentally documented
metacognitive-laziness pattern (Fan et al., 2025). Dialogic reliance
denotes iterative, conversational co-construction in which the writer
engages the GenAI tool as an interlocutor while retaining authorship of
the resulting synthesis, extending sociocultural accounts of learning
through interaction (Vygotsky, 1978) and social-cognitive models of
writing that position responsive others within the task environment
(Hayes, 1996).

The typology corresponds in informative ways to the four factors that
Hou et al. (2025) derived empirically in the problem-solving domain.
Their reflective use ~purposeful engagement combining prompt revision,
critical reading, and iterative refinement of AI output ~spans territory
the present typology differentiates: its orchestration component
parallels the Deliberate Use facet of Strategic reliance, and its
iterative-refinement component anticipates Dialogic reliance. Their
cautious use (recognizing and reacting to AI errors) parallels the
Critical Evaluation facet, and their thoughtless use (copying task
prompts, adopting output without major changes) parallels Dependent
reliance. Two features of this correspondence merit emphasis. First,
that an inductive analysis conducted in a different task domain,
country, and sample recovered factors aligning with a theory-derived
typology constitutes mutual corroboration of a kind that purely
inductive replication cannot supply convergence across derivation
logics, not merely across samples. Second, the correspondence is
structurally diagnostic: the two facets that Hou et
al.\textquotesingle s data separated into distinct factors (reflective
and cautious use) are precisely the two facets the present typology
subsumes under Strategic reliance, raising a question the confirmatory
analyses reported in Section 4 are designed to answer: whether strategic
reliance is better modeled as one factor or two. The remaining
non-overlap is symmetric and task-diagnostic: Hou et
al.\textquotesingle s collaborative use captures reliance mediated by
peers, a dimension their collaborative problem-solving context
foregrounds and individual academic writing does not, while Instrumental
reliance has no counterpart in their structure, plausibly because
surface-level delegation manifests in extended writing where editing,
rephrasing, and condensing are distinct, recurrent functions ~in ways
bounded group problem-solving does not afford. Each
instrument\textquotesingle s coverage thus reflects its task ecology;
the inclusion of Instrumental reliance extends the
construct\textquotesingle s coverage rather than relabeling it.

These four theoretically derived types, and the two dimensions
organizing them, informed the development of the GenAI-RTS items.

\subsection{Validity as argument: The Standards
framework}

The \emph{Standards for Educational and Psychological Testing} (AERA et
al., 2014) define validity as the degree to which evidence and theory
support the interpretations and uses of test scores, organized across
five sources of evidence: test content, response processes, internal
structure, relationships to other variables, and consequences of
testing. On the argument-based view this framework instantiates (Kane,
2013), validation is a structured case built across these sources rather
than a property certified by any single coefficient, with reliability a
necessary but not sufficient condition. Section 3.1 details how this
framework structures the present validation.

\begin{table*}[t]
\centering
\footnotesize
\caption*{\textbf{Table 1.} \emph{Published Instruments Measuring Student Engagement With GenAI}}
\begin{tabular}{@{} >{\raggedright\arraybackslash}p{(\textwidth - 12\tabcolsep) * \real{0.1549}} >{\raggedright\arraybackslash}p{(\textwidth - 12\tabcolsep) * \real{0.1656}} >{\raggedright\arraybackslash}p{(\textwidth - 12\tabcolsep) * \real{0.1496}} >{\raggedright\arraybackslash}p{(\textwidth - 12\tabcolsep) * \real{0.1549}} >{\raggedright\arraybackslash}p{(\textwidth - 12\tabcolsep) * \real{0.1870}} >{\raggedright\arraybackslash}p{(\textwidth - 12\tabcolsep) * \real{0.0962}} >{\raggedright\arraybackslash}p{(\textwidth - 12\tabcolsep) * \real{0.0919}}@{}}
\toprule\noalign{}
Instrument / study
 & Construct and task context
 & Development
 & Sample
 & Validity evidence and qualitative validation
 & Reliability
 & Invariance
 \\
\midrule\noalign{}

Reliance on GenAI scale (Hou et al., 2025) & Reliance behaviors in
problem-solving; two structured collaborative problem-solving tasks &
Inductive (EFA; factors named post hoc) & Singapore; 1 university; EFA n
= 800; CFAs n = 730 / 1,173 & Internal structure (EFA, 2 CFAs); limited
behavioral alignment; qualitative check partial (chat histories; high
scorers only) & $\alpha$ = .84; subscales $\geq$ .65 & No \\
AIDep-22 (Wu et al., 2026) & AI dependence as maladaptive disposition
(emotional, functional, cognitive, loss of control); general AI use
(non-task-specific) & Hypothesized dimensions; expert review + cognitive
interviews; structure via EFA/CFA & China; 1 university; two samples, N
= 400 each & Content; internal structure (EFA, CFA); convergent /
discriminant / criterion; qualitative work at item development only & $\alpha$
= .87 (total); subscales .86--.89 & Configural / metric across split
samples only \\
GAI tool-use scale (Barcelona \& Dela Cruz, 2025) & GenAI use practices
and perspectives (research, communication, reliance, ethics); general
academic use & Mixed (interviews of 20 students + scoping review;
inductive extraction) & Philippines; N = 793 (two pilots) & Internal
structure (EFA, CFA); internal consistency; convergent; qualitative work
at item generation only & Loadings .72--.93; $\alpha$ reported & No \\
SLA-GAI (Xia et al., 2025) & Student learning agency in GenAI-supported
contexts (adjacent construct); GenAI-supported learning generally &
Mixed (literature + expert interviews; EFA/CFA) & China; two stages; EFA
n = 268; second sample split EFA / CFA n = 268 / 245 & Internal
structure (EFA, CFA); qualitative work at item development only & Per
subscale & No \\
GenAI-RTS (this study) & Reliance as behavioral repertoire in academic
writing (4 types; 5 facets); academic writing across process stages &
Deductive (a priori theory; tested against rival 1--5-factor and
higher-order models) & U.S. Minority-Serving Institution; preliminary
survey n = 143; validation N = 382 (no racial majority) & Content;
response processes; internal structure (competing CFAs, invariance,
Rasch); relationships to other variables (Standards framework);
qualitative validation (14 interviews; maximum variation; all score
levels) & $\alpha$ = .74--.88; $\omega$ = .75--.88; AVE/CR & Yes --- scalar (gender,
first-gen, STEM) \\
\bottomrule
\end{tabular}
\par\vspace{2pt}
{\footnotesize \emph{\textbf{Note.}} AI literacy and attitude scales (e.g., Chung et al., 2025; Gümüş \& Kara, 2025; Marengo et al., 2025; Zhang et al., 2025) measure adjacent constructs and are excluded; see Section 2.2. EFA = exploratory factor analysis; CFA = confirmatory factor analysis; AVE = average variance extracted; CR = composite reliability.\par}
\end{table*}

\section{Methodology}

\subsection{Validity framework and
design}

Data were drawn from a larger sequential explanatory mixed-methods study
of undergraduate reliance on GenAI in academic writing. The present
article reports the components relevant to instrument validation. The
validation design was guided by the \emph{Standards} framework
introduced in Section 2.4. Each source of validity evidence was linked a
priori to specific data sources and analyses. Quantitative survey data
supported factor analytic, reliability, invariance, and Rasch analyses,
while interview data provided complementary evidence about students'
response processes. Table 2 presents the evidence-source-by-data-source
matrix governing the study. Two features of the design are particularly
important. First, the logic is theory-then-confirmation: the items were
specified deductively from named theoretical frameworks and screened
through faculty expert review, and the full confirmatory burden,
competing-model tests, invariance, and Rasch diagnostics were carried by
the validation sample. Second, the qualitative strand was designed to
contribute directly to validation rather than simply illustrate the
quantitative findings. Interview participants were selected through
maximum variation sampling across the full range of reliance scores.
This allowed the response-process analysis to examine how students at
different score levels interpreted and responded to the items.

Consistent with the argument-based view of validity, validation evidence
accumulates across a research program rather than within a single study
(Kane, 2013; Krupa et al., 2020; Sondergeld, 2020). The present study
examines four sources of validity evidence: test content, response
processes, internal structure, and relationships to other variables.
Evidence regarding the consequences of testing is deferred to future
research examining the instrument's use in advising and instructional
contexts, consistent with recent validation practice in education
(Mata-McMahon et al., 2023). None of the instruments reviewed in Section
2.2 has been validated using this framework. Applying it here provides a
coherent structure for evaluating the GenAI-RTS and makes clear both the
evidence examined in the present study and the evidence that remains to
be established.

\begin{table*}[t]
\centering
\footnotesize
\caption*{\textbf{Table 2.} \emph{Validity Evidence Sources and Associated Data Sources and Analyses}}
\begin{tabular}{@{} >{\raggedright\arraybackslash}p{(\textwidth - 2\tabcolsep) * \real{0.2778}} >{\raggedright\arraybackslash}p{(\textwidth - 2\tabcolsep) * \real{0.7222}}@{}}
\toprule\noalign{}
Validity evidence source (AERA et al., 2014)
 & Data sources and analyses in the present study
 \\
\midrule\noalign{}

Test content & Theoretical derivation of the four-type typology from
named frameworks (Section 2.3); literature-to-item mapping (Table 5);
faculty expert review; preliminary survey evidence informing construct
identification (\emph{n} = 143) \\
Response processes & Directed content analysis of 14 maximum-variation
interviews; coded alignment between scale-implied behavior and
participants' self-described behavior across high, mid, and low scorers
per subscale \\
Internal structure & Competing-model confirmatory factor analyses (one-
through five-factor and higher-order specifications); reliability ($\alpha$, $\omega$,
mean inter-item correlations); convergent/discriminant indices (AVE, CR,
HTMT); configural--metric--scalar measurement invariance (gender,
first-generation status, STEM/non-STEM); Rasch rating-scale and item-fit
analyses per subscale \\
Relationships to other variables & A priori convergent expectations
(Strategic reliance $\times$ AI literacy), discriminant expectations
(theoretically patterned subscale intercorrelations), and criterion
expectations (reliance-type differences across ten writing-process and
outcome variables) \\
Consequences of testing & Deferred to future research on the
instrument's use in advising and instructional contexts (rationale and
precedent in Section 3.1) \\
\bottomrule
\end{tabular}
\par\vspace{2pt}
{\footnotesize \emph{Note.} AERA = American Educational Research Association; AVE = average variance extracted; CR = composite reliability; HTMT = heterotrait--monotrait ratio.\par}
\end{table*}

\subsection{Participants and
setting}

The study was conducted at a public R1 doctoral university in the
mid-Atlantic United States with a federal Minority-Serving Institution
designation. The institutional context is integral to the validation
argument rather than incidental to it: published validations of related
instruments rest on demographically homogeneous, single-nation samples
(Section 2.2), and the present sample extends validation to a racially
diverse Minority-Serving Institution and examines measurement invariance
across gender, first-generation status, and academic discipline (STEM
vs. non-STEM majors). To our knowledge, these are the first reported
tests of scalar measurement invariance across demographic groups for an
instrument measuring student reliance on GenAI.

\textbf{Sample size justification.} The achieved validation sample
exceeded common recommendations for confirmatory factor analysis of an
instrument of this length: the validation sample of~\emph{N}~= 382
provides approximately 19 participants per item, within the widely cited
10--20-per-item guideline, and retains group sizes adequate for
two-group measurement-invariance testing (all groups $\geq$ 125 after
listwise deletion).

\textbf{Preliminary survey.} Instrument development was also informed by
a preliminary online survey of 143 undergraduates that examined GenAI
familiarity, use, perceived benefits and concerns, and open-ended
accounts of challenges in academic work. The survey findings were used
to identify context-specific expressions of theoretically defined
reliance patterns and to inform item wording rather than to derive the
factor structure.

\textbf{Main sample.} Participants were recruited over six weeks through
departmental listservs coordinated with faculty liaisons across more
than 30 academic departments, in-class announcements by cooperating
instructors, and institutional research participation channels.
Recruitment reached approximately 1,500 students, yielding an estimated
response rate of 26\%. Of 385 survey submissions, two platform-preview
test records and one case exceeding the a priori 20\% missing-data
threshold were excluded, yielding an analytic sample of~\emph{N}~= 382
with negligible item-level missingness. Table 3 presents sample
characteristics. Mean age was 20.8 years (\emph{SD}~= 3.9; range 17--56)
among the 291 participants who reported age. Participants reported a
mean self-reported GPA of 3.44 (\emph{SD}~= 0.53) and spent an average
of 7.9 hours per week (\emph{SD}~= 8.9) on academic writing. The sample
was racially and ethnically diverse: 34.3\% identified as White, 27.2\%
as Black or African American, 24.9\% as Asian, and 7.6\% as Hispanic or
Latino. In addition, 33.5\% were first-generation college students and
54.2\% were STEM majors.

\textbf{Interview subsample.} Fourteen survey respondents who expressed
willingness to participate in an interview were selected using maximum
variation purposive sampling (Patton, 2015) across four dimensions:
dominant reliance type, AI literacy level (tercile splits on the
composite score), first-generation status, and academic discipline.
Sampling across the full range of reliance scores, rather than selecting
only high scorers, allowed the response-process analysis to examine item
interpretation across different score levels (Table 2). Interviews
averaged 46 minutes (range 31--60 minutes). Thematic saturation
monitoring indicated that new codes ceased emerging by approximately the
tenth interview, with the final interviews confirming existing
categories.

\textbf{Ethics.} The study was approved by the
university\textquotesingle s Institutional Review Board (Protocol
\#03-07-24-1408). Survey participants provided digital informed consent,
and interview participants provided written consent with verbal
reconfirmation at the start of each session. Fifty randomly selected
survey participants received a \$10 incentive, and interview
participants received \$25. Survey responses were anonymous, and
interview participants were assigned pseudonyms.

\begin{table*}[t]
\centering
\footnotesize
\caption*{\textbf{Table 3.} \emph{Demographic Characteristics of the Main Sample (N = 382)}}
\begin{tabular}{@{} >{\raggedright\arraybackslash}p{(\textwidth - 4\tabcolsep) * \real{0.6333}} >{\raggedright\arraybackslash}p{(\textwidth - 4\tabcolsep) * \real{0.1833}} >{\raggedright\arraybackslash}p{(\textwidth - 4\tabcolsep) * \real{0.1833}}@{}}
\toprule\noalign{}
Characteristic
 & \emph{n}
 & \%
 \\
\midrule\noalign{}

Gender & & \\
Woman
 & 186 & 48.7 \\
Man
 & 163 & 42.7 \\
Non-binary, genderqueer, or transgender
 & 25 & 6.5 \\
Prefer not to answer
 & 8 & 2.1 \\
Race/ethnicity & & \\
White
 & 131 & 34.3 \\
Black or African American
 & 104 & 27.2 \\
Asian
 & 95 & 24.9 \\
Hispanic or Latino
 & 29 & 7.6 \\
Another identity (incl. Native Hawaiian/Pacific Islander)
 & 16 & 4.2 \\
Prefer not to answer
 & 7 & 1.8 \\
Academic year & & \\
Freshman (0--29 credits)
 & 100 & 26.2 \\
Sophomore (30--59 credits)
 & 83 & 21.7 \\
Junior (60--89 credits)
 & 76 & 19.9 \\
Senior ($\geq$90 credits)
 & 120 & 31.4 \\
Not reported
 & 3 & 0.8 \\
First-generation college student & & \\
Yes
 & 128 & 33.5 \\
No
 & 254 & 66.5 \\
Major/discipline & & \\
STEM
 & 207 & 54.2 \\
Social sciences
 & 80 & 20.9 \\
Arts \& humanities
 & 55 & 14.4 \\
Performing arts
 & 5 & 1.3 \\
Other
 & 35 & 9.2 \\
Family socioeconomic background & & \\
Working class
 & 66 & 17.3 \\
Lower-middle
 & 76 & 19.9 \\
Middle
 & 153 & 40.1 \\
Upper-middle
 & 71 & 18.6 \\
Affluent
 & 1 & 0.3 \\
Prefer not to answer
 & 15 & 3.9 \\
\bottomrule
\end{tabular}
\par\vspace{2pt}
{\footnotesize \emph{Note.} Age: \emph{M} = 20.8, \emph{SD} = 3.9 (\emph{n} = 291 reporting; eight birth-year entries converted to ages). Self-reported GPA: \emph{M} = 3.44, \emph{SD} = 0.53. Weekly hours on academic writing: \emph{M} = 7.9, \emph{SD} = 8.9. Percentages may not total 100 due to rounding.\par}
\end{table*}

\subsection{Instrument}

Instrument development proceeded in sequential phases: theoretical
construct specification, deductive item generation, faculty expert
review, refinement to the final 20-item form, and multi-source
validation in the present study. Table 4 summarizes the sequence.

\begin{table*}[t]
\centering
\footnotesize
\caption*{\textbf{Table 4.} \emph{Instrument Development Process}}
\begin{tabular}{@{} >{\raggedright\arraybackslash}p{(\textwidth - 4\tabcolsep) * \real{0.2457}} >{\raggedright\arraybackslash}p{(\textwidth - 4\tabcolsep) * \real{0.3846}} >{\raggedright\arraybackslash}p{(\textwidth - 4\tabcolsep) * \real{0.3697}}@{}}
\toprule\noalign{}
Stage
 & Purpose
 & Outcome
 \\
\midrule\noalign{}

Construct identification & Derive reliance types from theory (Section
2.3) and preliminary survey evidence (Section 3.2) & 4 constructs; 5
measurement facets \\
Item generation & Deductive item writing from named source frameworks &
Initial pool of 25 items \\
Faculty expert review & Content relevance, clarity, and redundancy
screening & 20 items retained \\
Validation (present study) & Multi-source psychometric evaluation &
Five-facet structure retained; one wording revision and a 5-point
response format recommended \\
\bottomrule
\end{tabular}
\end{table*}

\textbf{Construct identification.} The four reliance types were
specified before any item was written, derived from the theoretical
frameworks documented in Section 2.3 and informed by preliminary survey
evidence on undergraduates' GenAI use (Section 3.2). Strategic reliance
was specified with two facets, Deliberate Use and Critical Evaluation,
yielding five measurement facets in total.

\textbf{Item generation.} An initial pool of 25 items was generated
deductively from the five measurement facets. Each item was designed to
represent a single facet-level behavior grounded in its source
framework. Negative wording and double-barreled items were avoided. A
post hoc readability check of the final 20 items yielded a mean Flesch
Kincaid grade level of 11.2 and a Flesch Reading Ease score of 41.
Because readability was not formally assessed during item development,
these results are reported descriptively.

\textbf{Expert review.} The initial item pool was reviewed by three
faculty experts with expertise in quantitative methodology and
statistics, educational technology, and literacy education. Each expert
independently evaluated every item for clarity, alignment with its
intended reliance construct, representativeness of the construct domain,
and relevance to academic writing. Feedback was consolidated by the
research team, and items were revised or removed when concerns involved
redundancy, unclear wording, or weak alignment with the intended facet.
Of the 25 initial items, expert review led to rewording for clarity and
the removal of five items (three for content redundancy with retained
items and two for weak alignment with their intended construct),
yielding the final 20-item instrument. No formal content-validity
indices were computed, and no external panel beyond this faculty review
was convened, a constraint acknowledged in the limitations.

\textbf{Final instrument.} The GenAI-RTS contains 20 items representing
four theoretically derived reliance types and five measurement facets.
Strategic reliance comprises two facets, Deliberate Use and Critical
Evaluation, while Instrumental, Dependent, and Dialogic reliance are
each represented by one facet.

\textbf{GenAI Reliance Types Scale.} The GenAI-RTS comprises 20 items
distributed across the four theoretically derived reliance types, with
all items rated on 7-point Likert scales (1 = \emph{Strongly Disagree},
7 = \emph{Strongly Agree}). Strategic reliance is measured by eight
items operationalizing its two theoretical facets (Section 2.3):four
Deliberate Use items informed by metacognitive awareness theory and
measurement (Schraw \& Dennison, 1994) and four Critical Evaluation
items adapted from critical AI literacy competencies (Long \& Magerko,
2020). Instrumental reliance (four items) operationalizes
efficiency-targeted, surface-level delegation; Dependent reliance (four
items) operationalizes wholesale outsourcing and uncritical acceptance,
with items adapted from self-efficacy and learned-helplessness
measurement traditions (Bandura, 1997); and Dialogic reliance (four
items) operationalizes iterative co-construction, drawing on
sociocultural accounts of learning through interaction (Vygotsky, 1978).
Table 5 presents the literature-to-item mapping with a sample item per
facet; the full instrument appears in Appendix A. Subscale scores are
item means (range 1--7); the Strategic composite is the mean of its two
facet means, weighting the facets equally. All validation analyses
reported here use continuous subscale scores. For known-groups analyses,
students were classified by their highest subscale mean (yielding
Strategic n = 131, Instrumental n = 118, Dialogic n = 116, Dependent n =
17); no minimum-score threshold was applied. Interpretively, subscale
scores describe reliance profiles rather than assign students to types,
no diagnostic cut scores are proposed, and the highest-subscale
classification serves research description only (see Section 5).

\begin{table*}[t]
\centering
\footnotesize
\caption*{\textbf{Table 5.} \emph{Item Blueprint for the GenAI-RTS: Domains, Definitions, Theoretical Sources, and Sample Items}}
\begin{tabular}{@{} >{\raggedright\arraybackslash}p{(\textwidth - 6\tabcolsep) * \real{0.2030}} >{\raggedright\arraybackslash}p{(\textwidth - 6\tabcolsep) * \real{0.3579}} >{\raggedright\arraybackslash}p{(\textwidth - 6\tabcolsep) * \real{0.3472}} >{\raggedright\arraybackslash}p{(\textwidth - 6\tabcolsep) * \real{0.0919}}@{}}
\toprule\noalign{}

Construct facet & Definition and theoretical source & Sample item &
Items \\
Strategic, Deliberate Use & Planned, goal-directed engagement with AI.
Metacognitive regulation: planning and monitoring (Schraw \& Dennison,
1994) & ``I set clear objectives for using generative AI tools before
each writing session.'' & 4 \\
Strategic, Critical Evaluation & Verification and appraisal of AI output
before incorporation. Critical AI literacy: output evaluation
competencies (Long \& Magerko, 2020) & ``I verify facts in text produced
by generative AI tools against authoritative sources.'' & 4 \\
Instrumental & Efficiency-targeted delegation of discrete writing tasks.
Surface approach to learning (Biggs, 1987) & ``I use generative AI tools
to rephrase or clarify complex sentences in my drafts.'' & 4 \\
Dependent & Wholesale outsourcing with uncritical acceptance of AI
output. Over-reliance patterns adapted from self-efficacy and
learned-helplessness measurement (Bandura, 1997); cognitive offloading
(Risko \& Gilbert, 2016) & ``I accept outputs from generative AI tools
by default, even when I could compose those sections myself.'' & 4 \\
Dialogic & Iterative co-construction of ideas through interaction.
Sociocultural co-construction (Vygotsky, 1978) & ``I use generative AI
tools as brainstorming partners to explore new ideas or perspectives.''
& 4 \\
\bottomrule
\end{tabular}
\par\vspace{2pt}
{\footnotesize \emph{Note.} Full item text for all 20 items appears in Appendix A.\par}
\end{table*}

\textbf{Measures used as validity criteria.} Four sets of external
measures provided evidence based on relationships to other variables. AI
literacy was measured using six items informed by Allen and
Kendeou\textquotesingle s (2024) framework ($\alpha$ = .78); this six-item
measure served as the core AI literacy score. A separate seven-item
prior exposure index showed lower reliability ($\alpha$ = .66) and was
therefore interpreted with caution. An overall AI literacy composite was
computed as Expectancy-value beliefs were measured by five items adapted
from expectancy-value theory (Wigfield \& Eccles, 2000). The five items
formed a unidimensional composite ($\alpha$ = .92; one-factor CFA CFI = .98,
standardized loadings .78--.85), and were combined as a single
task-value index rather than modeled as separate expectancy-value
subconstructs. The items covered expectancy for success, perceived
efficiency (an AI-context extension of utility value capturing time and
effort savings; cf. Venkatesh et al., 2003), intrinsic enjoyment,
attainment value, and utility value. Writing process engagement was
measured using 12 items, with three items assessing AI use at each stage
of writing: planning ($\alpha$ = .89), drafting ($\alpha$ = .87), revising ($\alpha$ = .94),
and editing ($\alpha$ = .90). Perceived writing outcomes were measured by six
single-item indicators covering overall quality, self-efficacy, clarity,
grammar and style, originality, and critical thinking.

\textbf{Criterion definition.} A measurement clarification governs all
evidence based on relationships to other variables in Section 4.5 and
must be stated before any result is interpreted. The writing process and
outcome items were phrased as "I use generative AI tools to {[}achieve
X{]}." They therefore measure students\textquotesingle{} self-reported
use of AI for specific writing functions and the benefits they perceive
from that use. They do not measure independently assessed writing
quality or cognitive ability. Under this criterion, theoretically
coherent predictions differ from those an independent-quality criterion
would generate: students whose reliance pattern is defined by restraint
and verification (Strategic) should report lower AI-assisted attainment
precisely because their engagement involves withholding, whereas
students who outsource broadly (Dependent) should report high
AI-assisted attainment, including high AI-assisted self-efficacy. The a
priori expectations in Section 3.5 are stated against this criterion.

\subsection{Procedures}

\textbf{Survey administration.} The survey was administered through the
university\textquotesingle s Qualtrics platform. Item order was
randomized within construct blocks to reduce order effects. Key items
used forced response settings, while sensitive demographic questions
included a "prefer not to answer" option. The survey was optimized for
mobile devices and included a progress indicator. Median completion time
was 15 minutes (range 10--22 minutes among retained respondents). Before
analysis, exclusion criteria were specified for platform preview
records, failed attention checks, and more than 20\% missing data;
Qualtrics additionally flagged completion times below eight minutes
(approximately half the median) for manual review.

\textbf{Careless-response documentation.}~Beyond the a priori screening,
careless responding was documented following the longstring procedure of
Hou et al. (2025); flagged cases were treated as candidates for
sensitivity analysis rather than definitively classified as careless
responders, given the randomized item order. For each retained
respondent, the longstring index, the maximum number of consecutive
identical responses, was computed across the full survey battery of
items in presentation order (\emph{M}~= 10.47,~\emph{SD}~=
10.12,~\emph{Mdn}~= 6, range 2--56); because item order was randomized
within blocks, the index is conservative with respect to block
boundaries. This mirrors Hou et al.\textquotesingle s (2025) computation
of the index across all survey questions rather than only the target
scale. Twenty-eight cases (7.3\%) exceeded 1.5 times the interquartile
range above the third quartile (index \textgreater{} 26.5), including
four respondents who selected an identical response across the entire
survey; Qualtrics\textquotesingle{} native straightlining metric
corroborated this pattern. Rather than excluding these cases post hoc,
which would depart from the prespecified exclusion criteria, all
substantive analyses were accompanied by sensitivity analyses omitting
the flagged cases. All structural conclusions were robust: subscale
reliabilities shifted by at most $-$.033 and remained acceptable, the
retained measurement model\textquotesingle s fit was essentially
unchanged (CFI = .911, RMSEA = .076 vs. .916 and .079 in the full
sample), and discriminant-validity indices were stable or strengthened
(the Critical Evaluation--Dependent HTMT ratio remained well below the
.85 threshold, indicating clear discriminant separation). Sensitivity
results are noted alongside the corresponding analyses in Section 4.

\textbf{Interview procedures.} Semi-structured interviews followed a
protocol developed and refined through three preliminary interviews.
Interviews were conducted through institutionally licensed video
conferencing. For participants who preferred written interaction, a
structured asynchronous format with sequential follow-up probes was
used. Preliminary interviews suggested that both formats produced
comparable depth. Interviews were recorded with consent, transcribed,
and anonymized using pseudonyms. The protocol excerpt relevant to
reliance behaviors appears in Appendix B.

\subsection{Data analysis}

Analyses are reported in the order the Results section follows,
organized by validity evidence source. Statistical significance was
evaluated at $\alpha$ = .05 throughout, with effect sizes reported for all
group comparisons.

\textbf{Screening and distributional treatment}

Item-level missingness in the analytic sample was negligible (10 of
7,640 responses on the 20 GenAI-RTS items, 0.13\%; 56 of 21,392 across
the full 56-item battery, 0.26\%) and consistent with missing completely
at random by Little's (1988) test for both sets, $\chi^2$(130) = 98.77,
\emph{p} = .98, and $\chi^2$(1238) = 1226.58, \emph{p} = .59, respectively.
Analyses requiring complete cases used listwise deletion (e.g.,
confirmatory models, \emph{n} = 375); descriptive and correlational
analyses used pairwise deletion. Because the 7-point response format is
ordinal, the confirmatory analyses estimated under normal-theory maximum
likelihood (ML) were accompanied by diagonally weighted least squares
(DWLS) sensitivity estimations; treating responses with seven categories
as approximately continuous under ML is defensible (Rhemtulla et al.,
2012), and reporting both estimators establishes that structural
conclusions are not estimator-dependent.

\textbf{Structural evaluation strategy}

Because the items were developed deductively, structural evaluation
proceeded through competing confirmatory models rather than exploratory
factor analysis. This approach allowed the proposed structure to be
tested against theoretically plausible alternatives.

\textbf{Confirmatory factor analysis (competing models)}

Rather than fitting only the hypothesized structure, six competing
measurement models were specified a priori and compared in the main
sample: (M1) a single general factor; (M2) two factors separating
Strategic reliance from all engagement-oriented items; (M3) three
factors merging the Instrumental and Dialogic items, the alternative
implied by their expected intercorrelation; (M4) the hypothesized
four-factor model with Strategic as a single eight-item factor; (M5) a
five-factor model separating Strategic reliance into its Deliberate Use
and Critical Evaluation facets; and (M6) a higher-order model with
Strategic specified as a second-order factor over the two facets
alongside the three remaining first-order factors. Fit was evaluated
with $\chi^2$/df, CFI, TLI, RMSEA, and SRMR, interpreting CFI/TLI $\geq$ .90 and
RMSEA $\leq$ .08 as conventional acceptability thresholds while noting Hu and
Bentler's (1999) stricter guidelines and Marsh et al.'s (2004) cautions
against rigid universal cutoffs, with AIC and BIC for non-nested
comparison and DWLS sensitivity estimation as described above.

\textbf{Reliability and within-instrument convergent/discriminant
evidence}

Internal consistency was assessed with Cronbach's $\alpha$ and McDonald's $\omega$ per
subscale, with mean inter-item correlations evaluated against the
.20--.40 optimal range (Piedmont, 2014). Convergent and discriminant
adequacy within the instrument was assessed with average variance
extracted (AVE $\geq$ .50; Fornell \& Larcker, 1981), composite reliability
(CR $\geq$ .70), and heterotrait--monotrait ratios (HTMT; Henseler et al.,
2015), interpreting HTMT against both the strict .85 and liberal .90
criteria. One interpretive rule was fixed in advance: for the
Instrumental--Dialogic pair, whose intercorrelation was expected to be
the instrument's highest given the shared engagement-volume component
(Section 3.3), the structural test, whether the five-factor model
outperforms the three-factor model merging the pair (M5 vs. M3), was
designated the primary discriminant criterion, with HTMT reported
descriptively.

\textbf{Measurement invariance}

Multigroup confirmatory models tested configural, metric (equal
loadings), and scalar (equal loadings and intercepts) invariance of the
retained measurement model across three groupings: gender (women vs.
men; the non-binary, genderqueer, and transgender group, \emph{n} = 25,
was too small to model and is addressed in the limitations),
first-generation status, and STEM versus non-STEM major. Invariance at
each step was evaluated by $\Delta$CFI $\leq$ .01 and $\Delta$RMSEA $\leq$ .015 (Chen, 2007),
criteria less sample-size-sensitive than $\Delta$$\chi^2$. Group sizes (all $\geq$ 125 per
group after listwise deletion) are reported with each comparison;
two-group invariance testing at these \emph{n}s is adequately powered
for the loading- and intercept-level constraints evaluated.

\textbf{Rasch analysis}

Because Rasch models assume unidimensionality, Rasch analyses were
conducted within each of the five subscales identified by the
confirmatory analyses rather than across the full instrument. Andrich's
rating scale model evaluated whether the 7-point response format
functioned as intended, applying Linacre's (2002) criteria: at least 10
observations per category, monotonic advance of average person measures
across categories, ordered step calibrations advancing by at least 1.0
and less than 5.0 logits, and distinct probability-curve peaks for each
category. Threshold ordering was cross-validated with partial credit
model estimates obtained by marginal maximum likelihood, guarding
against estimation-method artifacts. Item and person fit were evaluated
with infit and outfit mean-squares (0.6--1.4; Wright \& Linacre, 1994),
and person and item reliability and separation were computed, with
Wright maps used to examine item targeting. A possibility was
anticipated explicitly at the design stage: if intermediate categories
of the 7-point format failed Linacre's criteria, category collapse would
be reported as a substantive finding with a concrete instrument-revision
recommendation, following the precedent of Mata-McMahon et al. (2023).

\textbf{Relationships to other variables}

Convergent, discriminant, and criterion expectations were specified a
priori against the AI-assisted-attainment criterion defined in Section
3.3 and summarized in Table 2: Strategic reliance was expected to
correlate positively with AI literacy; the Strategic--Dependent
intercorrelation was expected to be the instrument's lowest, reflecting
their conceptual opposition; Dependent reliance was expected to
correlate positively with AI-assisted writing self-efficacy; and
reliance types were expected to differentiate the ten writing-process
and outcome variables. Expectations were tested with Pearson
correlations and one-way analyses of variance with $\eta^2$ effect sizes,
using Tukey's HSD for homogeneous variances and Games--Howell otherwise.

\textbf{Response-process analysis}

Interview transcripts were analyzed by directed content analysis (Hsieh
\& Shannon, 2005) with the five measurement facets as the a priori
coding frame. For each facet, excerpts describing participants' actual
engagement behaviors were coded for alignment, partial alignment, or
divergence with the behavior the corresponding items imply, and the
analysis sampled participants across high, mid, and low score levels per
subscale, the maximum-variation design feature that distinguishes this
evidence from qualitative validation restricted to high scorers. Coding
was conducted in NVivo 15 with reflexive memoing and an audit trail;
divergences are reported alongside alignments, as divergence bounds
interpretation rather than invalidating it.

\textbf{Software}

Quantitative analyses were conducted in Python 3 (pandas, NumPy, and
SciPy), with confirmatory factor models estimated in semopy 2.3.11,
multigroup invariance models estimated by custom maximum-likelihood
routines validated to exact agreement with semopy single-group
estimates, and Rasch rating-scale models estimated by joint maximum
likelihood with threshold structure cross-validated against partial
credit estimates from girth (marginal maximum likelihood); Little's MCAR
test was computed in pyampute. Qualitative coding used NVivo 15.
Analysis code reproducing all reported statistics is available from the
authors.

\begin{quote}
\section{Results}
\end{quote}

Results are organized by the sources of validity evidence specified in
Table 2. Unless otherwise noted, analyses used the main sample
(\emph{N}~= 382), with~\emph{n}~= 375 for confirmatory models after
listwise deletion. Sensitivity analyses excluding the 28 cases flagged
by the longstring index are reported when they affect the interpretation
of a finding.

\subsection{Preliminary analyses}

Screening outcomes were as specified in Sections 3.2 and 3.4: of 385
submissions, three were excluded according to prespecified criteria: two
platform preview records and one case with 60.7\% missing data. This
yielded an analytic sample of~\emph{N}~= 382. Item-level missingness was
negligible (0.13\% on the 20 GenAI-RTS items; 0.26\% across the full
56-item battery) and was consistent with data missing completely at
random for both sets (Little\textquotesingle s test,~\emph{p}~= .98
and~\emph{p}~= .59, respectively). Table 6 presents descriptive
statistics, internal consistency estimates, and the correlation
structure for the five measurement facets. Table 7 presents item-level
statistics. Corrected item--total correlations, calculated within each
facet, ranged from .54 to .78 for 19 of the 20 items. The exception was
DEL3 (\emph{r}~= .28), consistent with the weaker factor loading
reported in Section 4.3.1. Two descriptive patterns merit note. First,
the two Strategic facets occupied different regions of the scale.
Critical Evaluation was the most strongly endorsed facet (\emph{M}~=
5.11,~\emph{SD}~= 1.38), whereas Deliberate Use fell below the scale
midpoint (\emph{M}~= 3.54,~\emph{SD}~= 1.29). Students, therefore,
endorsed Critical Evaluation more strongly than Deliberate Use, a
difference that would be obscured by the eight-item Strategic composite.
Second, the facet correlations showed substantial variation. Critical
Evaluation was nearly uncorrelated with Dependent reliance (\emph{r}~=
$-$.09), whereas Instrumental and Dialogic reliance showed the strongest
association among the five facets (\emph{r}~= .77).

\begin{table*}[t]
\centering
\footnotesize
\caption*{\textbf{Table 6.} \emph{Descriptive Statistics, Reliability, and Correlations Among the Five Measurement Facets (N = 382)}}
\begin{tabular}{@{} >{\raggedright\arraybackslash}p{(\textwidth - 18\tabcolsep) * \real{0.2297}} >{\raggedright\arraybackslash}p{(\textwidth - 18\tabcolsep) * \real{0.0780}} >{\raggedright\arraybackslash}p{(\textwidth - 18\tabcolsep) * \real{0.0780}} >{\raggedright\arraybackslash}p{(\textwidth - 18\tabcolsep) * \real{0.0801}} >{\raggedright\arraybackslash}p{(\textwidth - 18\tabcolsep) * \real{0.0801}} >{\raggedright\arraybackslash}p{(\textwidth - 18\tabcolsep) * \real{0.0908}} >{\raggedright\arraybackslash}p{(\textwidth - 18\tabcolsep) * \real{0.0908}} >{\raggedright\arraybackslash}p{(\textwidth - 18\tabcolsep) * \real{0.0908}} >{\raggedright\arraybackslash}p{(\textwidth - 18\tabcolsep) * \real{0.0908}} >{\raggedright\arraybackslash}p{(\textwidth - 18\tabcolsep) * \real{0.0908}}@{}}
\toprule\noalign{}
Facet
 & \emph{M}
 & \emph{SD}
 & Skew
 & Kurt
 & 1
 & 2
 & 3
 & 4
 & 5
 \\
\midrule\noalign{}

1. Deliberate Use & 3.54 & 1.29 & 0.19 & $-$0.26 & (.74) & .38 & .76 & .62
& .68 \\
2. Critical Evaluation & 5.11 & 1.38 & $-$0.86 & 0.57 & .30 & (.84) & .32
& .17 & .43 \\
3. Instrumental & 3.92 & 1.80 & $-$0.27 & $-$1.08 & .61 & .28 & (.88) & .61
& .89 \\
4. Dependent & 2.64 & 1.50 & 0.68 & $-$0.58 & .41 & $-$.09 & .53 & (.88) &
.65 \\
5. Dialogic & 4.29 & 1.70 & $-$0.51 & $-$0.67 & .54 & .37 & .77 & .55 &
(.86) \\
\bottomrule
\end{tabular}
\par\vspace{2pt}
{\footnotesize \emph{Note.} Pearson correlations below the diagonal; heterotrait--monotrait ratios (HTMT) above the diagonal, computed on absolute inter-item correlations (Henseler et al., 2015) using complete cases (\emph{n} = 375); Cronbach's $\alpha$ in parentheses on the diagonal. Skewness (Skew) and kurtosis (Kurt) are Fisher-adjusted; kurtosis is excess kurtosis. All response scales 1--7.\par}
\end{table*}

\begin{table*}[t]
\centering
\footnotesize
\caption*{\textbf{Table 7.} \emph{Item-Level Descriptive Statistics and Corrected Item--Total Correlations (N = 382)}}
\begin{tabular}{@{} >{\raggedright\arraybackslash}p{(\textwidth - 10\tabcolsep) * \real{0.1603}} >{\raggedright\arraybackslash}p{(\textwidth - 10\tabcolsep) * \real{0.1282}} >{\raggedright\arraybackslash}p{(\textwidth - 10\tabcolsep) * \real{0.1282}} >{\raggedright\arraybackslash}p{(\textwidth - 10\tabcolsep) * \real{0.1603}} >{\raggedright\arraybackslash}p{(\textwidth - 10\tabcolsep) * \real{0.1603}} >{\raggedright\arraybackslash}p{(\textwidth - 10\tabcolsep) * \real{0.2628}}@{}}
\toprule\noalign{}
Item
 & \emph{M}
 & \emph{SD}
 & Skewness
 & Kurtosis
 & Corrected item--total \emph{r}
 \\
\midrule\noalign{}

Deliberate Use & & & & & \\
DEL1
 & 3.78 & 1.93 & $-$0.08 & $-$1.28 & .64 \\
DEL2
 & 3.01 & 1.79 & 0.57 & $-$0.70 & .68 \\
DEL3
 & 4.83 & 1.60 & $-$0.57 & $-$0.17 & .28 \\
DEL4
 & 2.52 & 1.57 & 0.89 & $-$0.09 & .54 \\
Critical Evaluation & & & & & \\
CE1
 & 5.11 & 1.68 & $-$0.77 & $-$0.25 & .73 \\
CE2
 & 5.28 & 1.57 & $-$0.95 & 0.37 & .78 \\
CE3
 & 5.02 & 1.68 & $-$0.74 & $-$0.24 & .61 \\
CE4
 & 5.06 & 1.73 & $-$0.75 & $-$0.21 & .59 \\
Instrumental & & & & & \\
INST1
 & 3.81 & 2.12 & $-$0.08 & $-$1.47 & .70 \\
INST2
 & 3.98 & 2.13 & $-$0.17 & $-$1.41 & .75 \\
INST3
 & 4.06 & 2.15 & $-$0.19 & $-$1.44 & .74 \\
INST4
 & 3.82 & 2.02 & $-$0.12 & $-$1.33 & .76 \\
Dialogic & & & & & \\
DIAL1
 & 3.88 & 2.04 & $-$0.07 & $-$1.30 & .74 \\
DIAL2
 & 4.74 & 1.90 & $-$0.83 & $-$0.46 & .71 \\
DIAL3
 & 3.78 & 2.10 & $-$0.05 & $-$1.41 & .70 \\
DIAL4
 & 4.76 & 2.07 & $-$0.74 & $-$0.82 & .67 \\
Dependent & & & & & \\
DEP1
 & 2.79 & 1.75 & 0.69 & $-$0.62 & .72 \\
DEP2
 & 2.50 & 1.78 & 0.88 & $-$0.56 & .75 \\
DEP3
 & 2.76 & 1.79 & 0.66 & $-$0.82 & .78 \\
DEP4
 & 2.49 & 1.67 & 0.93 & $-$0.31 & .70 \\
\bottomrule
\end{tabular}
\par\vspace{2pt}
{\footnotesize \emph{Note.} Corrected item--total correlations are computed within facet (each item against the sum of the remaining three items of its facet). Skewness and kurtosis are Fisher-adjusted; kurtosis is excess kurtosis. Item codes correspond to Appendix A.\par}
\end{table*}

\subsection{Test content evidence}

Evidence based on test content comes from the deductive construct
development and faculty review described in Sections 2.3 and 3.3. The
item mapping showed that each theoretical facet was represented by four
items linked to a named source framework (Table 5): metacognitive
regulation for Deliberate Use, critical AI literacy competencies for
Critical Evaluation, surface approaches to learning for Instrumental
reliance, over-reliance and cognitive offloading for Dependent reliance,
and sociocultural co-construction for Dialogic reliance. All items were
written to reflect generative AI--supported academic writing behaviors,
with item wording informed by preliminary survey evidence on
undergraduates\textquotesingle{} GenAI use (\emph{n}~= 143; Section
3.2). Item content was reviewed by three faculty experts in quantitative
methodology and statistics, educational technology, and literacy
education (Section 3.3); no external panel was convened, and no formal
content-validity indices were computed, so content evidence rests on
theoretical traceability, preliminary survey grounding, and faculty
expert review, a constraint acknowledged in the limitations.

\subsection{Internal structure
evidence}

\subsubsection{Confirmatory factor analysis: Competing models}

Table 8 presents fit statistics for the six competing measurement
models. The model comparisons showed a consistent pattern across ML and
DWLS estimation. The four-type model (M4), which represented Strategic
reliance as a single eight-item factor, did not reach acceptable fit
under ML (CFI = .795, RMSEA = .122) and remained inadequate under DWLS
(RMSEA = .105). In contrast, the prespecified five-facet model (M5),
which modeled Deliberate Use and Critical Evaluation separately, showed
acceptable fit under ML, $\chi^2$(160) = 535.3, $\chi^2$/df = 3.35, CFI = .916, TLI
= .900, RMSEA = .079, SRMR = .088, and stronger fit under DWLS, CFI =
.976, TLI = .972, RMSEA = .066. M5 showed the strongest overall fit
among the competing models and the lowest AIC and BIC of the six
specifications.

The higher-order model (M6) tested whether Deliberate Use and Critical
Evaluation could be represented by a broader Strategic factor. This
model fell slightly below the acceptable ML criteria (CFI = .897, RMSEA
= .087) but showed acceptable fit under DWLS (CFI = .959). Two model
comparisons were particularly informative. First, combining Instrumental
and Dialogic reliance into one factor substantially worsened fit
relative to M5 ($\Delta$CFI = $-$.135 under ML), supporting their treatment as
distinct factors despite their strong correlation. Second, modeling
Deliberate Use and Critical Evaluation as separate facets substantially
improved fit over representing Strategic reliance as a single factor
($\Delta$CFI = +.121), supporting the five-facet measurement structure. Figure
2 displays the model comparisons.

Table 9 presents standardized loadings for the retained five-facet
model. Nineteen of the 20 items had standardized loadings of .66 or
higher (range = .66--.89). The exception was DEL3, ``I avoid using
generative AI tools unless their use aligns with my predetermined
writing objectives'' ($\lambda$ = .25). DEL3 was also the only item in the facet
framed in terms of restraint rather than engagement. Latent factor
correlations ranged from $-$.07 between Critical Evaluation and Dependent
reliance to .89 between Instrumental and Dialogic reliance. Excluding
the 28 cases flagged by the longstring index did not change the model
ordering or the overall interpretation of M5 fit (CFI = .911, RMSEA =
.076).

\begin{table*}[t]
\centering
\footnotesize
\caption*{\textbf{Table 8.} \emph{Fit of Six Competing Measurement Models (n = 375)}}
\begin{tabular}{@{} >{\raggedright\arraybackslash}p{(\textwidth - 20\tabcolsep) * \real{0.2094}} >{\raggedright\arraybackslash}p{(\textwidth - 20\tabcolsep) * \real{0.1068}} >{\raggedright\arraybackslash}p{(\textwidth - 20\tabcolsep) * \real{0.0662}} >{\raggedright\arraybackslash}p{(\textwidth - 20\tabcolsep) * \real{0.0598}} >{\raggedright\arraybackslash}p{(\textwidth - 20\tabcolsep) * \real{0.0598}} >{\raggedright\arraybackslash}p{(\textwidth - 20\tabcolsep) * \real{0.0908}} >{\raggedright\arraybackslash}p{(\textwidth - 20\tabcolsep) * \real{0.0684}} >{\raggedright\arraybackslash}p{(\textwidth - 20\tabcolsep) * \real{0.0940}} >{\raggedright\arraybackslash}p{(\textwidth - 20\tabcolsep) * \real{0.0940}} >{\raggedright\arraybackslash}p{(\textwidth - 20\tabcolsep) * \real{0.0598}} >{\raggedright\arraybackslash}p{(\textwidth - 20\tabcolsep) * \real{0.0908}}@{}}
\toprule\noalign{}
\multicolumn{9}{@{}l}{%
} & \multicolumn{2}{l@{}}{%
DWLS
} \\
\midrule\noalign{}

Model & $\chi^2$ (df) & $\chi^2$/df & CFI & TLI & RMSEA & SRMR & AIC & BIC & CFI &
RMSEA \\
M1: One factor & 1737.5 (170) & 10.22 & .648 & .607 & .157 & .138 &
27,689.2 & 27,924.8 & .913 & .123 \\
M2: Two factors (Strategic vs. engagement) & 1610.2 (169) & 9.53 & .676
& .636 & .151 & .136 & 27,563.9 & 27,803.5 & .918 & .119 \\
M3: Three factors (Instrumental + Dialogic merged) & 1143.3 (167) & 6.85
& .781 & .751 & .125 & .180 & 27,101.0 & 27,348.4 & .937 & .105 \\
M4: Four factors (hypothesized) & 1077.4 (164) & 6.57 & .795 & .762 &
.122 & .180 & 27,041.1 & 27,300.3 & .938 & .105 \\
\textbf{M5: Five factors (Strategic facets separated)} & \textbf{535.3
(160)} & \textbf{3.35} & \textbf{.916} & \textbf{.900} & \textbf{.079} &
\textbf{.088} & \textbf{26,507.0} & \textbf{26,781.9} & \textbf{.976} &
\textbf{.066} \\
M6: Higher-order (Strategic over two facets) & 620.4 (162) & 3.83 & .897
& .879 & .087 & .105 & 26,588.1 & 26,855.1 & .959 & .086 \\
\bottomrule
\end{tabular}
\par\vspace{2pt}
{\footnotesize \emph{Note.} ML estimates with DWLS sensitivity columns. DWLS indices run systematically higher than ML for these data; the model ordering is identical under both estimators, and M5 is the only model meeting acceptability criteria under ML. AIC and BIC computed from the ML log-likelihood with saturated means; lower values indicate better fit. Retained model in bold.\par}
\end{table*}

\begin{table*}[t]
\centering
\footnotesize
\caption*{\textbf{Table 9.} \emph{Standardized Factor Loadings, Five-Factor Model (n = 375)}}
\begin{tabular}{@{} >{\raggedright\arraybackslash}p{(\textwidth - 6\tabcolsep) * \real{0.4157}} >{\raggedright\arraybackslash}p{(\textwidth - 6\tabcolsep) * \real{0.0843}} >{\raggedright\arraybackslash}p{(\textwidth - 6\tabcolsep) * \real{0.4157}} >{\raggedright\arraybackslash}p{(\textwidth - 6\tabcolsep) * \real{0.0843}}@{}}
\toprule\noalign{}
Factor / item
 & \emph{$\lambda$}
 & Factor / item
 & \emph{$\lambda$}
 \\
\midrule\noalign{}

Deliberate Use & & Dependent & \\
DEL1 (set clear objectives)
 & .75 & DEP1 (accept outputs by default)
 & .80 \\
DEL2 (plan sections in advance)
 & .82 & DEP2 (uneasy starting without AI)
 & .82 \\
DEL3 (avoid unless aligned with objectives)
 & .25 & DEP3 (incorporate with minimal revision)
 & .85 \\
DEL4 (allocate dedicated time blocks)
 & .73 & DEP4 (rarely question accuracy)
 & .74 \\
Critical Evaluation & & Dialogic & \\
CE1 (verify facts against sources)
 & .81 & DIAL1 (multiple rounds of refinement)
 & .82 \\
CE2 (evaluate logical consistency)
 & .89 & DIAL2 (pose follow-up questions)
 & .78 \\
CE3 (identify biases/ethical issues)
 & .66 & DIAL3 (blend AI passages with own voice)
 & .77 \\
CE4 (revise extensively for rigor)
 & .69 & DIAL4 (brainstorming partner)
 & .76 \\
Instrumental & & & \\
INST1 (create outline/framework)
 & .77 & & \\
INST2 (rephrase/clarify sentences)
 & .81 & & \\
INST3 (condense source material)
 & .78 & & \\
INST4 (craft and refine prompts)
 & .85 & & \\
\bottomrule
\end{tabular}
\par\vspace{2pt}
{\footnotesize \emph{Note.} Item descriptors abbreviated; full wording in Appendix A. All loadings \emph{p} \textless{} .001.\par}
\end{table*}

\begin{figure*}[t]
\centering
\includegraphics[width=\textwidth]{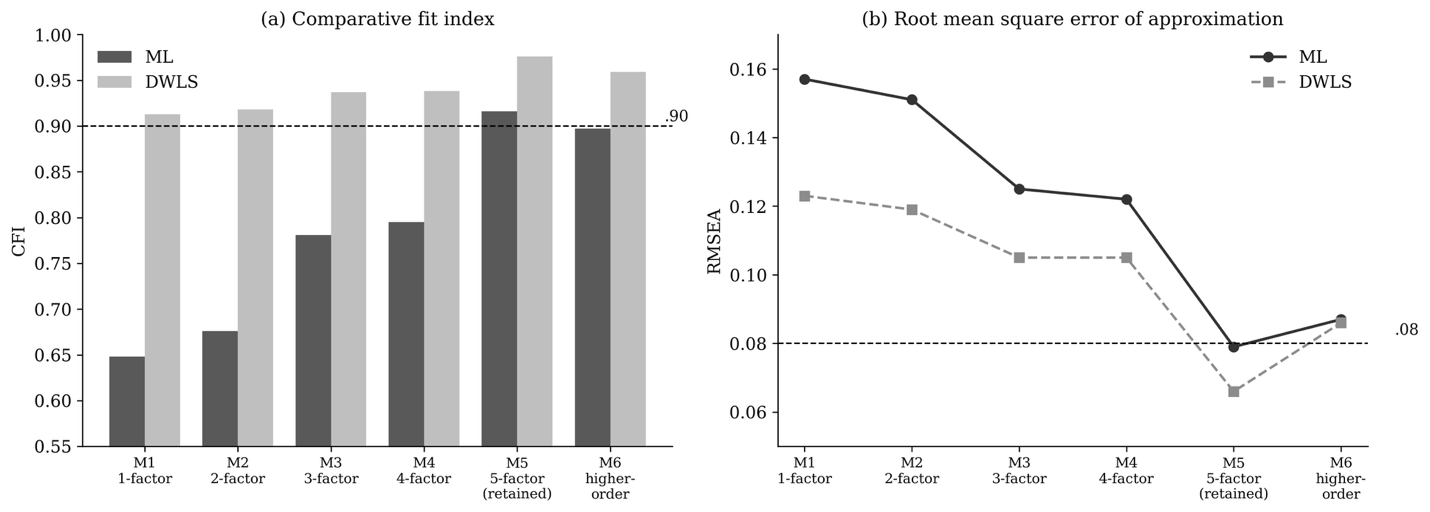}
\caption*{\textbf{Figure 2.} Fit of six competing measurement models under maximum likelihood (ML) and diagonally weighted least squares (DWLS). \emph{Note.} Dashed lines mark conventional acceptability criteria. The five-factor model (M5) is the only specification meeting ML criteria; the model ordering is identical under both estimators.}
\end{figure*}

\subsubsection{Reliability and convergent and discriminant evidence}

Table 10 presents reliability estimates and evidence of convergent and
discriminant validity. Omega and alpha were good for four facets ($\omega$ =
.85--.88) and acceptable for Deliberate Use ($\omega$ = .75), where the weak
third item depresses internal consistency. Average variance extracted
exceeded .50 for all facets except Deliberate Use (.46), and composite
reliabilities all exceeded .70. Mean inter-item correlations exceeded
the .20--.40 optimal range for four of five facets (.58--.65), an
expected property of short, homogeneous four-item subscales that is
reported for transparency; it indicates item redundancy within facets
rather than unreliability and bounds the breadth of each
facet\textquotesingle s content coverage. The HTMT matrix (Table 6,
above diagonal) localized discriminant pressure exactly where the design
anticipated it: nine of ten ratios fell below the strict .85 criterion,
with the Instrumental--Dialogic pair at .89, below the liberal .90
criterion but above the strict one. Under the interpretive rule fixed in
advance (Section 3.5), the structural test governs this pair, and the
M5-versus-M3 comparison in Table 8 resolves the issue: collapsing the
two factors is untenable. At the opposite pole, Critical Evaluation and
Dependent reliance were empirically near-orthogonal (HTMT = .17;
latent~\emph{r}~= $-$.07), the instrument\textquotesingle s clearest
discriminant evidence and a direct empirical signature of the
theoretical opposition between vigilant evaluation and wholesale
outsourcing. This separation strengthened in the careless-response
sensitivity analysis.

\begin{table*}[t]
\centering
\footnotesize
\caption*{\textbf{Table 10.} \emph{Reliability and Convergent Validity Indices by Facet (N = 382)}}
\begin{tabular}{@{} >{\raggedright\arraybackslash}p{(\textwidth - 12\tabcolsep) * \real{0.3117}} >{\raggedright\arraybackslash}p{(\textwidth - 12\tabcolsep) * \real{0.0909}} >{\raggedright\arraybackslash}p{(\textwidth - 12\tabcolsep) * \real{0.0909}} >{\raggedright\arraybackslash}p{(\textwidth - 12\tabcolsep) * \real{0.1039}} >{\raggedright\arraybackslash}p{(\textwidth - 12\tabcolsep) * \real{0.1039}} >{\raggedright\arraybackslash}p{(\textwidth - 12\tabcolsep) * \real{0.0909}} >{\raggedright\arraybackslash}p{(\textwidth - 12\tabcolsep) * \real{0.2078}}@{}}
\toprule\noalign{}

Facet & $\alpha$ & $\omega$ & MIC & AVE & CR & Loading range \\
Deliberate Use & .74 & .75 & .40 & .46 & .75 & .25--.82 \\
Critical Evaluation & .84 & .85 & .58 & .59 & .85 & .66--.89 \\
Instrumental & .88 & .88 & .64 & .64 & .88 & .77--.85 \\
Dependent & .88 & .88 & .65 & .65 & .88 & .74--.85 \\
Dialogic & .86 & .86 & .61 & .61 & .86 & .76--.82 \\
\bottomrule
\end{tabular}
\par\vspace{2pt}
{\footnotesize \emph{Note.} MIC = mean inter-item correlation; AVE = average variance extracted; CR = composite reliability (congeneric, equal to $\omega$ here). Loading ranges, $\omega$, AVE, and CR derive from the five-factor model estimated on complete cases (\emph{n} = 375); $\alpha$ and MIC use all available cases. The eight-item Strategic composite, modeled as a single factor, yields $\omega$ = .76 and AVE = .33, corroborating that it does not function as one dimension.\par}
\end{table*}

\subsubsection{Measurement invariance}

Table 11 presents configural, metric, and scalar invariance of the
five-factor model across gender, first-generation status, and STEM
versus non-STEM major. Configural fit sat at the adequate--borderline
boundary under ML in each grouping (CFI = .896--.905), consistent with
the full-sample ML solution; invariance itself is evaluated by the
change statistics, which were uniformly and decisively within criteria:
across all three groupings and both constraint steps, $\Delta$CFI never
exceeded .002 (criterion $\leq$ .01) and $\Delta$RMSEA never exceeded zero in the
adverse direction (criterion $\leq$ .015), with RMSEA improving at every
step. Scalar invariance, therefore, held for all three groupings,
licensing latent and observed mean comparisons across gender,
first-generation status, and disciplinary groups. To our knowledge, this
is the first reported evidence of scalar measurement invariance across
demographic groups for an instrument measuring student reliance on
generative AI, the precondition for the group-difference research
questions that motivate reliance measurement.

\begin{table*}[t]
\centering
\footnotesize
\caption*{\textbf{Table 11.} \emph{Measurement Invariance of the Five-Factor Model}}
\begin{tabular}{@{} >{\raggedright\arraybackslash}p{(\textwidth - 12\tabcolsep) * \real{0.2943}} >{\raggedright\arraybackslash}p{(\textwidth - 12\tabcolsep) * \real{0.1555}} >{\raggedright\arraybackslash}p{(\textwidth - 12\tabcolsep) * \real{0.0957}} >{\raggedright\arraybackslash}p{(\textwidth - 12\tabcolsep) * \real{0.1077}} >{\raggedright\arraybackslash}p{(\textwidth - 12\tabcolsep) * \real{0.0957}} >{\raggedright\arraybackslash}p{(\textwidth - 12\tabcolsep) * \real{0.1376}} >{\raggedright\arraybackslash}p{(\textwidth - 12\tabcolsep) * \real{0.1136}}@{}}
\toprule\noalign{}
Grouping / model
 & $\chi^2$ (df)
 & CFI
 & RMSEA
 & $\Delta$CFI
 & $\Delta$RMSEA
 & Decision
 \\
\midrule\noalign{}

Gender (women \emph{n} = 181, men \emph{n} = 161) & & & & & & \\
Configural
 & 691.1 (320) & .905 & .082 & --- & --- & --- \\
Metric
 & 706.0 (335) & .905 & .080 & .000 & $-$.002 & Hold \\
Scalar
 & 724.4 (350) & .904 & .079 & .001 & $-$.001 & Hold \\
First-generation (no \emph{n} = 250, yes \emph{n} = 125) & & & & & & \\
Configural
 & 772.0 (320) & .898 & .087 & --- & --- & --- \\
Metric
 & 794.4 (335) & .897 & .086 & .002 & $-$.001 & Hold \\
Scalar
 & 818.9 (350) & .895 & .085 & .002 & $-$.001 & Hold \\
STEM (yes \emph{n} = 203, no \emph{n} = 172) & & & & & & \\
Configural
 & 770.9 (320) & .900 & .087 & --- & --- & --- \\
Metric
 & 782.7 (335) & .900 & .084 & $-$.001 & $-$.002 & Hold \\
Scalar
 & 802.8 (350) & .899 & .083 & .001 & $-$.001 & Hold \\
\bottomrule
\end{tabular}
\par\vspace{2pt}
{\footnotesize \emph{Note.} Group \emph{n}s after listwise deletion. Criteria: $\Delta$CFI $\leq$ .01 and $\Delta$RMSEA $\leq$ .015 (Chen, 2007). $\Delta$ values are computed from unrounded fit indices and may differ slightly from differences between the rounded values shown; negative values indicate improvement under the added constraints. The non-binary, genderqueer, and transgender group (\emph{n} = 25) was too small for multigroup modeling; see limitations.\par}
\end{table*}

\subsubsection{Rasch analysis: Rating-scale functioning}

Rasch rating-scale analyses within each facet evaluated whether the
7-point response format functioned as intended. All seven response
categories met Linacre\textquotesingle s (2002) minimum frequency
criterion, with every category attracting at least 32 observations in
every facet. The format nevertheless failed the threshold-ordering
criteria in a consistent, interpretable pattern replicated across two
estimation methods (rating-scale JML and partial-credit MML): the step
calibrations adjacent to category 3 ("Slightly Disagree") were
disordered in 18 of the 20 items, indicating a category that is never
the most probable response at any level of the latent trait, and
category 3 was used less frequently than both adjacent categories in
every facet (e.g., 58 of 1,527 Dialogic responses). A secondary
disordering in the mild-agreement region, the step between categories 4
and 5 exceeding the subsequent step, appeared in 6 of the 20 items.
Together, these results indicate that respondents are not distinguishing
"Slightly Disagree" from its neighbors and that the disagreement side of
the scale carries more categories than respondents use.~

Item-difficulty orderings were stable across both estimators and
substantively interpretable: within Deliberate Use, aligning use with
predetermined objectives was the easiest behavior to endorse, and
allocating dedicated time blocks was the hardest, while the four
Critical Evaluation items clustered tightly in difficulty; evaluation
behaviors travel together; planning behaviors stratify. The
category-functioning verdict yields the instrument-revision
recommendation developed in Section 5: a 5-point format that merges the
two intermediate disagreement categories and the adjacent mild-agreement
region should function more cleanly, paralleling the category-collapse
finding of Mata-McMahon et al. (2023) for a 5-point frequency scale.
Figure 3 displays the category-usage profile underlying this diagnosis.

\begin{figure*}[t]
\centering
\includegraphics[width=\textwidth]{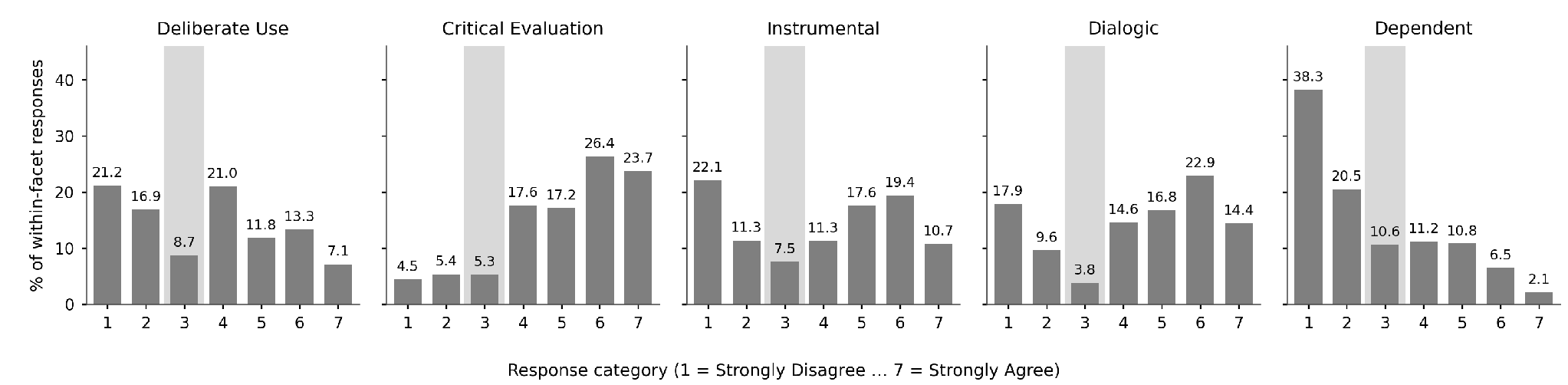}
\caption*{\textbf{Figure 3.} Response-category usage by facet (percentage of within-facet responses; model-free counts, N = 382). The shaded region marks category 3 (``Slightly Disagree''), used less frequently than both adjacent categories in every facet and the locus of the threshold disordering replicated across two estimation methods.}
\end{figure*}

\subsection{Response processes
evidence}

Directed content analysis of the 14 maximum-variation interviews
examined whether participants' descriptions of their AI use matched the
behaviors represented by the scale items. This provides evidence about
the link between item responses and score interpretation that factor
analysis alone cannot establish. Because participants were sampled
across a broad range of scores, the analysis examined response processes
among low, middle, and high scorers. This extends qualitative validation
beyond the high-scoring respondents emphasized in previous research (Hou
et al., 2025). Table 12 summarizes alignment evidence for each facet and
provides illustrative excerpts using participant pseudonyms.

Evidence of alignment was found for all five facets. Five participants
described selective restraint and boundary setting consistent with
Deliberate Use. Critical Evaluation appeared in accounts of actively
questioning and verifying AI output. Tyler, for example, explained, ``I
expect it to be wrong, and I'm kind of trying to disprove it.'' Nine
participants described bounded, task-specific delegation consistent with
Instrumental reliance. Seven described experiences of wholesale
outsourcing consistent with Dependent reliance, although these accounts
were retrospective, as discussed below. Eight described sustained,
iterative exchanges with AI consistent with Dialogic reliance. Sandra,
for example, described AI as ``a thinking partner'' used through
repeated and refining questions. Alignment was also evident across score
levels. Participants with low Dependent scores described careful
evaluation and verification practices consistent with limited dependent
reliance, while participants with mid-range Instrumental scores
described bounded, task-specific delegation consistent with their
scores.

Three findings also identified important limits to score interpretation.
First, no participant fit a single reliance category. All 14 described
multiple reliance patterns that varied across tasks and situations.
Scores should therefore be interpreted as profiles of co-occurring
reliance behaviors rather than as fixed or mutually exclusive student
types. This finding supports continuous subscale scoring and cautions
against categorical labeling.

Second, participants described Dependent reliance only as a past
behavior; none reported current uncritical use. This pattern may reflect
social desirability, changes in reliance over time, or both. The finding
raises the possibility that current self-reports may underestimate
dependent reliance, particularly among students reluctant to acknowledge
behaviors viewed negatively.

Third, one participant reported limiting AI use because of concern
following a disputed academic integrity accusation rather than because
of a deliberate learning strategy. The scale may record these forms of
restraint similarly even though they arise from different reasons. This
finding suggests that similar Deliberate Use scores may arise from
different motivations. The scale captures the reported behavior but may
not distinguish intentional learning strategies from restraint driven by
fear of institutional consequences. This limitation should be considered
when scores are used in policy-sensitive settings.

\begin{table*}[t]
\centering
\footnotesize
\caption*{\textbf{Table 12.} \emph{Response-Process Alignment: Facet $\times$ Interview Evidence (n = 14)}}
\begin{tabular}{@{} >{\raggedright\arraybackslash}p{(\textwidth - 6\tabcolsep) * \real{0.1335}} >{\raggedright\arraybackslash}p{(\textwidth - 6\tabcolsep) * \real{0.2618}} >{\raggedright\arraybackslash}p{(\textwidth - 6\tabcolsep) * \real{0.3205}} >{\raggedright\arraybackslash}p{(\textwidth - 6\tabcolsep) * \real{0.2842}}@{}}
\toprule\noalign{}

Facet & Aligned behavioral evidence (prevalence) & Illustrative excerpt
& Divergence / interpretive bound \\
Deliberate Use & Deliberate withholding and self-regulated
boundary-setting (5/14) & ``I try to avoid using AI. I only use AI as a
last resort\ldots'' (Tyler) & Externally induced restraint can mimic
deliberate restraint (1 case); low scores admit two mechanisms \\
Critical Evaluation & Adversarial verification; trust calibration
against authoritative sources & ``At this point I expect it to be wrong
and I'm kind of trying to disprove it\ldots{} So, I'll go back to the
textbook.'' (Tyler) & None observed; closest item--behavior
correspondence in the corpus \\
Instrumental & Task-directed, bounded problem-solving with retained
authorship (9/14) & ``I use the AI for the simple reason\ldots{} to get
things done, to solve problems that I face.'' (Elias) & Feedback-seeking
variants (AI as evaluator) sit at the facet's boundary with Dialogic
use \\
Dependent & High-volume, low-oversight outsourcing (7/14) & ``I used
to\ldots{} use AI to do almost all of my assignments and then eventually
I started to truly realize that I wasn't learning anything, so I
stopped.'' (Destiny) & Narrated only in past tense; concurrent
self-report may underestimate active dependency \\
Dialogic & Iterative prompting and co-construction with retained
synthesis (8/14) & ``I use AI as a thinking partner\ldots{} If I'm
stumped with how I should approach something\ldots{} I ask AI a
question.'' (Sandra) & Overlaps situationally with Instrumental use;
repertoires co-occur within persons \\
\bottomrule
\end{tabular}
\par\vspace{2pt}
{\footnotesize \emph{Note.} Prevalence = interviewees contributing at least one coded instance. All names are pseudonyms. Coding frame, procedures, and audit trail in Section 3.5.\par}
\end{table*}

\subsection{Relationships to other variables\textquotesingle{}
evidence}

Table 13 compares the observed findings with the a priori expectations
specified in Section 3.5 against the AI-assisted-attainment criterion
defined in Section 3.3. These expectations were based on external
measures of students' reported use of AI for writing functions and
perceived benefits rather than independently assessed writing quality.
All a priori expectations were supported. Strategic reliance was
positively associated with the AI literacy composite (r = .61), core AI
literacy (r = .51), and prior AI exposure (r = .55). Among the four
broad reliance type scores, the Strategic composite showed its weakest
association with Dependent reliance (r = .19), while the remaining
reliance types showed stronger positive associations with one another (r
= .53--.77). The four dominant profile groups differed significantly
across all ten writing process and perceived benefit variables, with all
omnibus tests reaching p \textless{} .001. Effect sizes ranged from $\eta^2$ =
.166 for grammar and style to $\eta^2$ = .329 for overall quality. Dependent
reliance was strongly associated with self-reported AI-assisted writing
self-efficacy (r = .64). This direction was consistent with the
prespecified expectation because the criterion reflects perceived
capability when using AI rather than independent writing self-efficacy.
Additional evidence for distinguishing Instrumental and Dialogic
reliance came from the dominant profile comparisons. Despite their
strong correlation, the Instrumental and Dialogic groups differed in
perceived originality (d = 0.45) and critical thinking (d = 0.41),
providing further support for retaining the two constructs separately.

\begin{table*}[t]
\centering
\footnotesize
\caption*{\textbf{Table 13.} \emph{A Priori Expectations and Observed Evidence (Relationships to Other Variables)}}
\begin{tabular}{@{} >{\raggedright\arraybackslash}p{(\textwidth - 4\tabcolsep) * \real{0.4274}} >{\raggedright\arraybackslash}p{(\textwidth - 4\tabcolsep) * \real{0.3269}} >{\raggedright\arraybackslash}p{(\textwidth - 4\tabcolsep) * \real{0.2457}}@{}}
\toprule\noalign{}

Expectation (stated in advance) & Observed & Verdict \\
Convergent: Strategic reliance correlates positively with AI literacy
(composite, core, exposure) & r = .61 / .51 / .55, all p \textless{}
.001 & Supported \\
Discriminant: Strategic--Dependent is the lowest subscale
intercorrelation (conceptual opposition) & r = .19; lowest of all pairs
& Supported \\
Discriminant pattern: high-engagement types (Instrumental, Dialogic,
Dependent) intercorrelate more strongly with one another than with
Strategic restraint & r = .53--.77 within triad & Supported \\
Criterion: dominant reliance type differentiates all ten writing-process
and outcome variables & All F p \textless{} .001; $\eta^2$ = .166--.329
(uniformly large) & Supported \\
Criterion (attainment frame): Dependent reliance correlates positively
with AI-assisted writing self-efficacy & r = .64, p \textless{} .001 &
Supported \\
Differentiation of the most-correlated pair: dominant-Instrumental vs.
dominant-Dialogic students differ on originality and critical thinking &
d = 0.45 (originality); d = 0.41 (critical thinking) & Supported \\
\bottomrule
\end{tabular}
\par\vspace{2pt}
{\footnotesize \emph{Note.} Group \emph{n}s for type comparisons: Strategic 131, Instrumental 118, Dialogic 116, Dependent 17; classification by highest subscale mean. Expectations specified in Sections 3.3 and 3.5 prior to analysis.\par}
\end{table*}

\section{Discussion}

This discussion addresses the central question of an instrument
validation study: What does the accumulated evidence show about the
quality of the GenAI-RTS as a measure? The discussion evaluates the
overall validity argument, explains the structural refinements supported
by the data, and positions the instrument within the emerging
measurement literature. Claims about AI reliance are considered only
when they inform the interpretation and use of GenAI-RTS scores.

\subsection{Summary of the validity
argument}

This study examined whether the GenAI-RTS demonstrates acceptable
evidence across four of the five validity sources specified in the
Standards. Table 14 summarizes the evidence for each claim. Evidence
based on test content came from the theory-driven development of the
five facets (Table 5), the literature-to-item mapping, and review by
three faculty experts. However, the absence of an independent external
panel and formal content validity indices remains the main limitation of
this evidence source.

Response process evidence showed alignment between participants'
accounts and the behaviors represented by all five facets across a broad
range of scores. The interviews also identified three limits to score
interpretation: scores represent overlapping profiles rather than fixed
types; current Dependent reliance may be underreported; and similar
Deliberate Use scores may arise from different motivations.

Internal structure evidence provided the strongest support for a
five-facet measurement structure, which outperformed five competing
models across both estimators. The five facets showed acceptable to good
reliability and scalar invariance across the three tested groupings.

Evidence based on relationships with other variables supported all
prespecified expectations.

The resulting argument is not that the instrument is flawless. The
evidence identifies clear limitations, including the weak third
Deliberate Use item, possible overlap within some facets, and problems
with the seven-point response format. At the same time, the evidence
supports specific score interpretations while clearly identifying where
those interpretations should be limited. This is the purpose of a
validity argument rather than a simple validity verdict (Kane, 2013).

\begin{table*}[t]
\centering
\footnotesize
\caption*{\textbf{Table 14.} \emph{The Validity Argument in Summary: Claims, Evidence, and Verdicts}}
\begin{tabular}{@{} >{\raggedright\arraybackslash}p{(\textwidth - 4\tabcolsep) * \real{0.3312}} >{\raggedright\arraybackslash}p{(\textwidth - 4\tabcolsep) * \real{0.4017}} >{\raggedright\arraybackslash}p{(\textwidth - 4\tabcolsep) * \real{0.2671}}@{}}
\toprule\noalign{}
Claim about GenAI-RTS scores
 & Principal evidence
 & Verdict
 \\
\midrule\noalign{}

Items represent the theorized reliance construct (content) &
Framework-derived facets; literature-to-item mapping; faculty expert
review; no external panel & Supported, with the expert-review gap
acknowledged \\
Responses reflect the behaviors the items imply (response processes) &
Item--behavior alignment across all facets and score levels in 14
maximum-variation interviews; three bounded divergences & Supported,
with three explicit interpretive bounds \\
Scores reflect five distinct facets organized by a four-type taxonomy
(internal structure) & M5 superior to five rivals under ML and DWLS; $\omega$ =
.75--.88; scalar invariance across gender, first-generation status, STEM
& Supported; four-type taxonomy retained at the theoretical level, five
facets at the measurement level \\
The 7-point format functions as intended (internal structure) &
Replicated threshold disordering at category 3; category-3 underuse
across facets & Not supported; 5-point revision recommended \\
Scores relate to external variables as theory predicts (relationships to
other variables) & Pre-stated expectations supported (convergent r =
.51--.61; $\eta^2$ = .166--.329; d = 0.41--0.45 for the most-correlated pair)
& Supported \\
Scores support consequential use decisions (consequences) & Deferred to
future research (Krupa et al., 2020; Sondergeld, 2020) & Not yet
evaluated \\
\bottomrule
\end{tabular}
\par\vspace{2pt}
{\footnotesize \emph{Note.} Verdicts correspond to the evidence reported in Section 4; bounds and gaps are constitutive of the argument, not appendices to it.\par}
\end{table*}

\subsection{The four-type taxonomy and the five-facet structure:
What held and what the data
refined}

The findings largely supported the proposed theoretical structure while
refining how Strategic reliance is represented at the measurement level.
At the theoretical level, the four broad reliance patterns remained
meaningful. Among the four broad type scores, Strategic and Dependent
reliance showed the weakest association (r = .19), while Critical
Evaluation and Dependent reliance were nearly unrelated at the facet
level.

What the data refined: Strategic reliance is one construct at the
theoretical level but two at the measurement level. Deliberate
orchestration and critical evaluation are dissociable behaviors;
students in this sample evaluated far more than they planned (M = 5.11
vs. 3.54), and forcing them into a single factor is the difference
between an inadequate and an acceptable model.

Three independent signals converge on this refinement: the
competing-model result, the facet asymmetry in the descriptives, and the
external replication discussed below. The instrument's most correlated
pair, Instrumental and Dialogic reliance (r = .77; HTMT = .89), was
interrogated from three directions and survived all three: structurally,
merging them degrades fit severely ($\Delta$CFI = $-$.135); theoretically, the
correlation reflects shared engagement volume, not shared construct ---
both patterns involve frequent AI contact while differing in its
character; and criterially, students dominant in each pattern differ on
exactly the outcomes the distinction predicts, originality (d = 0.45)
and critical thinking (d = 0.41).

Two instrument revisions follow directly from the evidence. The third
Deliberate Use item, the facet's only restraint-phrased item ($\lambda$ = .25),
should be reworded in the engagement-phrased direction of its facet or
replaced. And the response format should move from seven categories to
five: respondents demonstrably do not distinguish ``Slightly Disagree''
from its neighbors, a diagnosis invisible to factor-analytic methods and
recoverable only at the rating-scale level, paralleling Mata-McMahon et
al.'s (2023) category-collapse finding and illustrating why the
dual-tradition design carries its analytic cost. The recommended 5-point
format also converges with the response format Hou et al. (2025)
themselves adopted, independent designs arriving at the same
calibration.

\subsection{Position in the cumulative
literature}

The distinction between Deliberate Use and Critical Evaluation also has
a parallel in previous research. Hou et al. (2025) distinguished
reflective use from cautious use, a distinction that parallels the
separation of Deliberate Use and Critical Evaluation in the present
study.

When deductive and inductive derivations, applied to different
populations and tasks, fracture a construct at the same joint. This
convergence suggests that regulating AI engagement and evaluating AI
output may represent related but distinct competencies across more than
one task context. Further replication is needed to determine how broadly
this distinction generalizes.

The present study extends this work by testing a theory-derived
structure against competing models and adding forms of evidence not
previously reported in this literature, including measurement invariance
and an argument-based validation framework. The study also extends
validation to a racially diverse Minority Serving Institution, although
racial invariance could not be tested with the present sample sizes.

The presence of an Instrumental facet in this study, but not in Hou et
al.'s problem-solving structure, may reflect differences between task
contexts. Extended writing provides repeated opportunities for
surface-level delegation, such as polishing, rephrasing, and condensing,
that may be less central in bounded problem-solving tasks. This suggests
that some features of reliance may be shaped by the task context. The
reverse difference is also informative. Hou et al.'s collaborative use
factor, which involves reliance through peers, has no direct counterpart
in the GenAI-RTS. Together, these differences suggest that reliance
structures may partly depend on the tasks and social contexts in which
AI use occurs.

\subsection{Implications}

For researchers, the practical importance of scalar invariance is
straightforward. Comparisons across gender, first-generation status, or
academic discipline are meaningful only if respondents interpret the
items in the same way. To our knowledge, scalar invariance across
demographic groups had not previously been established for an instrument
measuring student reliance on GenAI. Researchers using the GenAI-RTS can
interpret observed differences across the tested groups as reflecting
differences in reliance rather than differences in how participants
understood the items.

For educators and practitioners, three implications follow. First,
scores should be interpreted as reliance profiles rather than fixed
student types. The response-process evidence showed that students used
different patterns of AI reliance depending on the writing task and
context. Therefore, interventions should focus on strengthening
productive reliance patterns rather than labeling students.

Second, high scores on the Dependent dimension provide the clearest
indicator of students who may benefit from AI literacy instruction.
Because Dependent reliance represents a pattern distinct from Critical
Evaluation, interventions should focus on strengthening students'
ability to critically evaluate AI-generated content rather than simply
reducing AI use.

Third, the response-process evidence carries a caution for
policy-sensitive uses: low engagement scores can reflect principled
strategy or fear of institutional consequences, and the instrument
cannot distinguish them; the GenAI-RTS is a formative diagnostic, not a
conduct-screening device.

\subsection{Limitations and future
directions}

Limitations should be considered, each with a corresponding direction
for future research. First, the structural evidence is based on a single
validation sample. Although competing CFA models were compared, the
five-factor structure should be replicated in an independent sample to
establish cross-validation. Second, the study was conducted at a single
institution using a cross-sectional design. Although interview data
suggested movement from dependent toward strategic reliance, these
retrospective accounts cannot establish developmental change.
Longitudinal studies across multiple institutions are needed. Third, all
measures are self-report: the study lacks the behavioral data such as AI
interaction logs with which Hou et al. (2025) anchored their qualitative
check, and triangulating GenAI-RTS scores against interaction-log
indicators would provide the strongest available test of the
response-process claims.

Fourth, evidence based on test content would be strengthened through
review by an independent panel of experts. Future administrations should
include formal content validation, such as expert ratings of item
relevance and clarity. Fifth, the criterion variables focused on
AI-assisted writing behaviors rather than independently assessed writing
quality. Future studies should examine whether reliance profiles predict
external outcomes such as writing quality, course performance, and
transfer of learning. These outcomes would also provide evidence
regarding the consequences of testing (Krupa et al., 2020; Sondergeld,
2020). Sixth, measurement invariance could only be examined across
two-group comparisons because the non-binary, genderqueer, and
transgender subgroup was too small for stable estimation. Multi-site
studies with more diverse samples would permit more comprehensive and
intersectional invariance testing.

In addition, the observed mean inter-item correlations suggest that
future versions of the scale could broaden the content represented
within each facet. Finally, the temporal stability of the GenAI-RTS has
not yet been established. Future studies should examine test--retest
reliability across appropriate time intervals.

\section{Conclusion}

The Generative AI Reliance Types Scale measures how undergraduates rely
on generative AI in academic writing, addressing a question that has
become increasingly important as generative AI use has become nearly
universal. Using quantitative and qualitative evidence from a
Minority-Serving Institution, the study provided validity evidence
across four of the five sources identified in the Standards for
Educational and Psychological Testing. Evidence supported the
instrument's theoretical foundation, response processes, internal
structure, measurement invariance across key student groups, and
relationships with external variables.

The instrument is intended for formative use, helping educators identify
reliance profiles and students who may benefit from AI literacy support,
rather than for academic misconduct screening. The findings also
identify two priorities for the next version of the instrument: revising
one item and adopting a five-point response scale.

Future research should evaluate the revised instrument across different
institutions, educational contexts, and writing tasks, ultimately,
against behavioral evidence of students' writing processes. By moving
beyond a simple distinction between AI use and non-use, the GenAI-RTS
enables researchers and educators to better understand how students
engage with AI during academic writing and to design more targeted AI
literacy interventions.

\section*{Data availability}

The de-identified item-level dataset and the analysis code reproducing
all reported statistics are available on request from the corresponding
author.

\section*{CRediT authorship contribution statement}

Shahin Hossain: Conceptualization, Methodology, Investigation, Formal
analysis, Data curation, Writing -- original draft, Writing -- review \&
editing. Tukhbita Afroz Nawmi: Validation, Writing -- review \& editing.

\section*{Funding}

This research did not receive any specific grant from funding agencies
in the public, commercial, or not-for-profit sectors.

\section*{Declaration of competing interest}

The authors declare no competing interests.

\section*{Declaration of generative AI and AI-assisted technologies in the writing process}

During the preparation of this work, the authors used Claude (Anthropic;
Fable 5, Opus 4.8, and Sonnet 4.6) and ChatGPT (OpenAI; GPT-5.5) to
improve language clarity, edit style, and format tables. After using
these tools, the authors reviewed and edited the content as needed and
take full responsibility for the content of the published article.

\section*{Appendix A. The Generative AI Reliance Types Scale (GenAI-RTS)}

Instructions: Please indicate your agreement with each statement using
the following 7-point scale: 1 = \emph{Strongly Disagree}; 2 =
\emph{Disagree}; 3 = \emph{Slightly Disagree}; 4 = \emph{Neither Agree
nor Disagree}; 5 = \emph{Slightly Agree}; 6 = \emph{Agree}; 7 =
\emph{Strongly Agree}.

\textbf{Strategic Reliance, Deliberate Use (DEL)}

1. I set clear objectives for using generative AI tools before each
writing session.

2. I plan in advance which sections of my assignment I will draft with
generative AI tools.

3. I avoid using generative AI tools unless their use aligns with my
predetermined writing objectives.

4. I allocate dedicated time blocks to interact with generative AI tools
during my writing process.

\textbf{Strategic Reliance, Critical Evaluation (CE)}

5. I verify facts in text produced by generative AI tools against
authoritative sources.

6. I evaluate the logical consistency and coherence of text produced by
generative AI tools.

7. I identify potential biases or ethical issues in text produced by
generative AI tools.

8. I revise text produced by generative AI tools extensively to ensure
academic rigor.

\textbf{Instrumental Reliance (INST)}

9. I use generative AI tools to create an outline or structural
framework for my writing before I begin drafting.

10. I use generative AI tools to rephrase or clarify complex sentences
in my drafts.

11. I use generative AI tools to condense lengthy source material into
concise summaries.

12. I craft and refine prompts to guide generative AI tools toward
specific writing objectives.

\textbf{Dependent Reliance (DEP)}

13. I accept outputs from generative AI tools by default, even when I
could compose those sections myself.

14. I feel uneasy starting a writing task without first consulting
generative AI tools for guidance.

15. I incorporate suggestions from generative AI tools into my drafts
with minimal revision.

16. I rarely question the accuracy or reliability of suggestions
provided by generative AI tools.

\textbf{Dialogic Reliance (DIAL)}

17. I engage in multiple rounds of prompts and refinements with
generative AI tools to co-author text.

18. I pose follow-up questions to generative AI tools to deepen or
nuance their responses.

19. I blend passages from generative AI tools with my own writing voice
within the same draft.

20. I use generative AI tools as brainstorming partners to explore new
ideas or perspectives.

\emph{Scoring:} subscale scores are item means (1--7); the Strategic
composite is the mean of the Deliberate Use and Critical Evaluation
facet means. \emph{Recommended revisions for the next version (Section
5.2):} reword item 3 in the engagement-phrased direction of its facet;
administer with a 5-point response format.

\textbf{Appendix B. Interview Protocol Excerpt (Reliance-Relevant
Sections)}

\emph{Note. The protocol is reproduced as administered; interview
questions used the term ``LLM,'' whereas survey items used ``generative
AI tools.''}

The semi-structured protocol comprised 13 core questions in four
sections; the two sections most relevant to the response-process
analysis are excerpted below. The full protocol is available from the
author.

\textbf{Section 1: LLM usage contexts and experiences}

\textbullet\  Can you walk me through a recent writing assignment where you used an
LLM? What was the assignment, how did you use the LLM, and what was the
outcome?

\textbullet\  Thinking across your different courses and writing assignments, how
does your LLM use vary depending on the context?

\textbullet\  Can you describe a time when using an LLM went particularly well? What
made it successful? Can you describe a time when it did not go well?
What went wrong?

\textbf{Section 2: Reliance type understanding and self-perception}

\textbullet\  Based on your survey responses, you were classified as exhibiting
primarily {[}strategic/instrumental/dependent/dialogic{]} reliance. Does
this characterization resonate with how you think about your LLM use?
Why or why not?

\textbullet\  How do you distinguish between using an LLM as a helpful tool versus
relying on it too much? Where do you draw that line for yourself?

\textbullet\  Some students describe GenAI as ``thinking partners,'' others as
``productivity tools,'' and others as ``crutches.'' Which metaphor best
captures your experience, and why?

\section*{References}
{\small
\begin{hangparas}{1.2em}{1}
Allen, L. K., \& Kendeou, P. (2024). ED-AI Lit: An interdisciplinary framework for AI literacy in education. \emph{Policy Insights from the Behavioral and Brain Sciences, 11}(1), 3--10. https://doi.org/10.1177/23727322231220339

American Educational Research Association, American Psychological Association, \& National Council on Measurement in Education. (2014). \emph{Standards for educational and psychological testing}. American Educational Research Association.

Bandura, A. (1997). \emph{Self-efficacy: The exercise of control}. W. H. Freeman.

Barcelona, A., \& Dela Cruz, S. R. (2025). Development and validation of a scale measuring students' use of generative artificial intelligence tools. \emph{International Journal of Evaluation and Research in Education, 14}(5), 3612--3621. https://doi.org/10.11591/ijere.v14i5.34809

Bereiter, C., \& Scardamalia, M. (1987). \emph{The psychology of written composition}. Lawrence Erlbaum Associates.

Biggs, J. B. (1987). \emph{Student approaches to learning and studying}. Australian Council for Educational Research.

Biggs, J. B. (1999). \emph{Teaching for quality learning at university}. Society for Research into Higher Education \& Open University Press.

Bulathwela, S., Pérez-Ortiz, M., Holloway, C., Cukurova, M., \& Shawe-Taylor, J. (2024). Artificial intelligence alone will not democratise education: On educational inequality, techno-solutionism and inclusive tools. \emph{Sustainability, 16}(2), 781. https://doi.org/10.3390/su16020781

Chen, F. F. (2007). Sensitivity of goodness of fit indexes to lack of measurement invariance. \emph{Structural Equation Modeling, 14}(3), 464--504. https://doi.org/10.1080/10705510701301834

Chung, K., Kim, S., Jang, Y., Choi, S., \& Kim, H. (2025). Developing an AI literacy diagnostic tool for elementary school students. \emph{Education and Information Technologies, 30}, 1013--1044. https://doi.org/10.1007/s10639-024-13097-w

Cotton, D. R. E., Cotton, P. A., \& Shipway, J. R. (2024). Chatting and cheating: Ensuring academic integrity in the era of ChatGPT. \emph{Innovations in Education and Teaching International, 61}(2), 228--239. https://doi.org/10.1080/14703297.2023.2190148

Fan, Y., Tang, L., Le, H., Shen, K., Tan, S., Zhao, Y., Shen, Y., Li, X., \& Gašević, D. (2025). Beware of metacognitive laziness: Effects of generative artificial intelligence on learning motivation, processes, and performance. \emph{British Journal of Educational Technology, 56}(2), 489--530. https://doi.org/10.1111/bjet.13544

Fornell, C., \& Larcker, D. F. (1981). Evaluating structural equation models with unobservable variables and measurement error. \emph{Journal of Marketing Research, 18}(1), 39--50. https://doi.org/10.1177/002224378101800104

Freeman, J. (2025). \emph{Student generative AI survey 2025} (HEPI Policy Note 61). Higher Education Policy Institute.

Gerlich, M. (2025). AI tools in society: Impacts on cognitive offloading and the future of critical thinking. \emph{Societies, 15}(1), 6. https://doi.org/10.3390/soc15010006

Graham, S., \& Harris, K. R. (2000). The role of self-regulation and transcription skills in writing and writing development. \emph{Educational Psychologist, 35}(1), 3--12. https://doi.org/10.1207/S15326985EP3501\_2

Gümüş, M. M., \& Kara, M. (2025). Development and validation of the Generative AI Literacy for Learning Scale (GenAI-LLs). \emph{Australasian Journal of Educational Technology, 41}(4), 1--16.

Hayes, J. R. (1996). A new framework for understanding cognition and affect in writing. In C. M. Levy \& S. Ransdell (Eds.), \emph{The science of writing: Theories, methods, individual differences, and applications} (pp. 1--27). Lawrence Erlbaum Associates.

Henseler, J., Ringle, C. M., \& Sarstedt, M. (2015). A new criterion for assessing discriminant validity in variance-based structural equation modeling. \emph{Journal of the Academy of Marketing Science, 43}(1), 115--135. https://doi.org/10.1007/s11747-014-0403-8

Hou, C., Zhu, G., Sudarshan, V., Lim, F. S., \& Ong, Y. S. (2025). Measuring undergraduate students' reliance on Generative AI during problem-solving: Scale development and validation. \emph{Computers \& Education, 234}, Article 105329. https://doi.org/10.1016/j.compedu.2025.105329

Hsieh, H.-F., \& Shannon, S. E. (2005). Three approaches to qualitative content analysis. \emph{Qualitative Health Research, 15}(9), 1277--1288. https://doi.org/10.1177/1049732305276687

Hu, K. (2023, February 2). ChatGPT sets record for fastest-growing user base. \emph{Reuters}. https://www.reuters.com/technology/chatgpt-sets-record-fastest-growing-user-base-analyst-note-2023-02-01/

Hu, L., \& Bentler, P. M. (1999). Cutoff criteria for fit indexes in covariance structure analysis: Conventional criteria versus new alternatives. \emph{Structural Equation Modeling, 6}(1), 1--55. https://doi.org/10.1080/10705519909540118

Kane, M. T. (2013). Validating the interpretations and uses of test scores. \emph{Journal of Educational Measurement, 50}(1), 1--73. https://doi.org/10.1111/jedm.12000

Kellogg, R. T. (2008). Training writing skills: A cognitive developmental perspective. \emph{Journal of Writing Research, 1}(1), 1--26.

Kosmyna, N., Hauptmann, E., Yuan, Y. T., Situ, J., Liao, X. H., Beresnitzky, A. V., Braunstein, I., \& Maes, P. (2025). \emph{Your brain on ChatGPT: Accumulation of cognitive debt when using an AI assistant for essay writing task} {[}Preprint{]}. arXiv. https://arxiv.org/abs/2506.08872

Krupa, E., Bostic, J., \& Shih, J. (2020). Validation in mathematics education: An introduction. In J. Bostic, E. Krupa, \& J. Shih (Eds.), \emph{Quantitative measures of mathematical knowledge: Researching instruments and perspectives} (pp. 1--13). Routledge.

Lee, J. D., \& See, K. A. (2004). Trust in automation: Designing for appropriate reliance. \emph{Human Factors, 46}(1), 50--80. https://doi.org/10.1518/hfes.46.1.50\_30392

Linacre, J. M. (2002). Optimizing rating scale category effectiveness. \emph{Journal of Applied Measurement, 3}(1), 85--106.

Little, R. J. A. (1988). A test of missing completely at random for multivariate data with missing values. \emph{Journal of the American Statistical Association, 83}(404), 1198--1202. https://doi.org/10.1080/01621459.1988.10478722

Liu, Z. M., Hwang, G. J., Chen, C. Q., Chen, X. D., \& Ye, X. D. (2024). Integrating large language models into EFL writing instruction: Effects on performance, self-regulated learning strategies, and motivation. \emph{Computer Assisted Language Learning, 39}(3), 466--490. https://doi.org/10.1080/09588221.2024.2389923

Lodge, J. M., \& Loble, L. (2026). \emph{Artificial intelligence, cognitive offloading and implications for education}. University of Technology Sydney. https://doi.org/10.71741/4pyxmbnjaq.31302475

Logg, J. M., Minson, J. A., \& Moore, D. A. (2019). Algorithm appreciation: People prefer algorithmic to human judgment. \emph{Organizational Behavior and Human Decision Processes, 151}, 90--103. https://doi.org/10.1016/j.obhdp.2018.12.005

Long, D., \& Magerko, B. (2020). What is AI literacy? Competencies and design considerations. In \emph{Proceedings of the 2020 CHI Conference on Human Factors in Computing Systems} (pp. 1--16). ACM. https://doi.org/10.1145/3313831.3376727

Marengo, A., Karaoglan-Yilmaz, F. G., Yılmaz, R., \& Ceylan, M. (2025). Development and validation of generative artificial intelligence attitude scale for students. \emph{Frontiers in Computer Science, 7}, Article 1528455. https://doi.org/10.3389/fcomp.2025.1528455

Marsh, H. W., Hau, K.-T., \& Wen, Z. (2004). In search of golden rules: Comment on hypothesis-testing approaches to setting cutoff values for fit indexes and dangers in overgeneralizing Hu and Bentler's (1999) findings. \emph{Structural Equation Modeling, 11}(3), 320--341. \url{https://doi.org/10.1207/s15328007sem1103_2}

Mata-McMahon, J., Haslip, M. J., \& Kruse, L. (2023). Validation study of the Early Childhood Educators\textquotesingle{} Spiritual Practices in the Classroom (ECE-SPC) instrument using Rasch.~\emph{International Journal of Children\textquotesingle s Spirituality, 28}(3--4), 99--132.~\url{https://doi.org/10.1080/1364436X.2023.2218590}

Mogavi, R. H., Deng, C., Kim, J. J., Zhou, P., Kwon, Y. D., Metwally, A. H. S., Tlili, A., Bassanelli, S., Bucchiarone, A., Gujar, S., Nacke, L. E., \& Hui, P. (2024). ChatGPT in education: A blessing or a curse? A qualitative study exploring early adopters' utilization and perceptions. \emph{Computers in Human Behavior: Artificial Humans, 2}(1), Article 100027. https://doi.org/10.1016/j.chbah.2023.100027

Parasuraman, R., \& Riley, V. (1997). Humans and automation: Use, misuse, disuse, abuse. \emph{Human Factors, 39}(2), 230--253. https://doi.org/10.1518/001872097778543886

Patton, M. Q. (2015). \emph{Qualitative research and evaluation methods} (4th ed.). SAGE.

Piedmont, R. L. (2014). Inter-item correlations. In A. C. Michalos (Ed.), \emph{Encyclopedia of quality of life and well-being research} (pp. 3303--3304). Springer. https://doi.org/10.1007/978-94-007-0753-5\_1493

\begin{quote} information technology: Toward a unified view.~\emph{MIS Quarterly, 27}(3), 425--478.~\url{https://doi.org/10.2307/30036540} \end{quote}

Rhemtulla, M., Brosseau-Liard, P. É., \& Savalei, V. (2012). When can categorical variables be treated as continuous? A comparison of robust continuous and categorical SEM estimation methods under suboptimal conditions. \emph{Psychological Methods, 17}(3), 354--373. https://doi.org/10.1037/a0029315

Risko, E. F., \& Gilbert, S. J. (2016). Cognitive offloading. \emph{Trends in Cognitive Sciences, 20}(9), 676--688. https://doi.org/10.1016/j.tics.2016.07.002

Rudolph, J., Tan, S., \& Tan, S. (2023). ChatGPT: Bullshit spewer or the end of traditional assessments in higher education? \emph{Journal of Applied Learning and Teaching, 6}(1), 342--363. https://doi.org/10.37074/jalt.2023.6.1.9

Schemmer, M., Kühl, N., Benz, C., Bartos, A., \& Satzger, G. (2023). Appropriate reliance on AI advice: Conceptualization and the effect of explanations. In \emph{Proceedings of the 28th International Conference on Intelligent User Interfaces} (pp. 410--422). ACM. https://doi.org/10.1145/3581641.3584066

Schraw, G., \& Dennison, R. S. (1994). Assessing metacognitive awareness. \emph{Contemporary Educational Psychology, 19}(4), 460--475. https://doi.org/10.1006/ceps.1994.1033

Sondergeld, T. A. (2020). Shifting sights on STEM education quantitative instrumentation development: The importance of moving validity evidence to the forefront rather than a footnote. \emph{School Science and Mathematics, 120}(5), 259--261. https://doi.org/10.1111/ssm.12410

Stephenson, R., \& Armstrong, C. (2026). \emph{Student generative AI survey 2026} (HEPI Report 199). Higher Education Policy Institute.

Venkatesh, V., Morris, M. G., Davis, G. B., \& Davis, F. D. (2003). User acceptance of

Vygotsky, L. S. (1978). \emph{Mind in society: The development of higher psychological processes}. Harvard University Press.

Wigfield, A., \& Eccles, J. S. (2000). Expectancy--value theory of achievement motivation. \emph{Contemporary Educational Psychology, 25}(1), 68--81. https://doi.org/10.1006/ceps.1999.1015

Wright, B. D., \& Linacre, J. M. (1994). Reasonable mean-square fit values. \emph{Rasch Measurement Transactions, 8}(3), 370.

Wu, H., Ni, H., Luo, W., \& Wu, T. (2026). Development and validation of the AI dependence scale for Chinese undergraduates and a preliminary exploration. \emph{Frontiers in Psychology, 16}, Article 1725393. https://doi.org/10.3389/fpsyg.2025.1725393

Xia, L., Shen, K., Sun, H., An, X., \& Dong, Y. (2025). Developing and validating the student learning agency scale in generative artificial intelligence (AI)-supported contexts. \emph{Education and Information Technologies, 30}(10), 13999--14021. https://doi.org/10.1007/s10639-024-13137-5

Zhang, H., Perry, A., \& Lee, I. (2025). Developing and validating the artificial intelligence literacy concept inventory: An instrument to assess artificial intelligence literacy among middle school students. \emph{International Journal of Artificial Intelligence in Education, 35}(1), 398--438. https://doi.org/10.1007/s40593-024-00398-x
\end{hangparas}
}

\end{document}